\documentclass[letterpaper]{article} 
\usepackage{aaai2026}  
\usepackage{amsmath} 
\usepackage{booktabs} 
\usepackage{amssymb}
\usepackage{times}  
\usepackage{helvet}  
\usepackage{courier}  
\usepackage[hyphens]{url}  
\usepackage{graphicx} 
\usepackage{natbib}  
\usepackage{caption} 
\usepackage{algorithm}
\usepackage{algpseudocode}
\usepackage{xcolor, multirow, colortbl}
\usepackage{pifont}

\newcommand{\best}[1]{\textbf{#1}}
\newcommand{\second}[1]{\underline{#1}}
\definecolor{mygrey}{cmyk}{0, 0, 0, 0.5}
\newcommand{\smallgrey}[1]{\textcolor{mygrey}}   

\usepackage{newfloat}
\usepackage{listings}
\DeclareCaptionStyle{ruled}{labelfont=normalfont,labelsep=colon,strut=off} 
\floatstyle{ruled}
\newfloat{listing}{tb}{lst}{}
\floatname{listing}{Listing}
\title{Training and Inference within 1 Second -- Tackle Cross-Sensor Degradation of Real-World Pansharpening with Efficient Residual Feature Tailoring}

\author{
    Tianyu Xin\textsuperscript{\rm 1}\equalcontrib,
    Jin-Liang Xiao\textsuperscript{\rm 1}\equalcontrib,
    Zeyu Xia\textsuperscript{\rm 1}\equalcontrib,
    Shan Yin\textsuperscript{\rm 1},
    Liang-Jian Deng\textsuperscript{\rm 1}\thanks{Corresponding author.}
}
\affiliations{
    \textsuperscript{\rm 1}University of Electronic Science and Technology of China\\

    tyxin@std.uestc.edu.cn, jinliang\_xiao@163.com, zeyuxia42@std.uestc.edu.cn, yins@std.uestc.edu.cn, liangjian.deng@uestc.edu.cn
}

\usepackage{bibentry}

\begin{document}

\maketitle

\begin{abstract}
Deep learning methods for pansharpening have advanced rapidly, yet models pretrained on data from a specific sensor often generalize poorly to data from other sensors.
Existing methods to tackle such cross-sensor degradation include retraining model or zero-shot methods, but they are highly time-consuming or even need extra training data. 
To address these challenges, our method first performs modular decomposition on deep learning-based pansharpening models, revealing a general yet critical interface where high-dimensional fused features begin mapping to the channel space of the final image. 
A Feature Tailor is then integrated at this interface to address cross-sensor degradation at the feature level, and is trained efficiently with physics-aware unsupervised losses. Moreover, our method operates in a patch-wise manner, training on partial patches and performing parallel inference on all patches to boost efficiency.
Our method offers two key advantages: (1) \textit{Improved Generalization Ability}: it significantly enhance performance in cross-sensor cases. (2) \textit{Low Generalization Cost}: it achieves sub-second training and inference, requiring only partial test inputs and no external data, whereas prior methods often take minutes or even hours.
Experiments on the real-world data from multiple datasets demonstrate that our method achieves state-of-the-art quality and efficiency in tackling cross-sensor degradation. For example, training and inference of $512\times512\times8$ image within \textit{0.2 seconds} and $4000\times4000\times8$ image within \textit{3 seconds} at the fastest setting on a commonly used RTX 3090 GPU, which is over 100 times faster than zero-shot methods. 
\end{abstract}

\section{Introduction} \label{sec:intro}
Remote sensing images are widely utilized in various fields. Due to hardware constraints, satellites typically capture low-resolution multispectral (LRMS) images and high-resolution panchromatic (PAN) images separately. And the target of pansharpening is to fuse an LRMS image and a PAN image to a high-resolution multispectral (HRMS) image that preserves both spectral and spatial information.

\begin{figure}[t]
  \centering
  \includegraphics[width=0.48\textwidth]{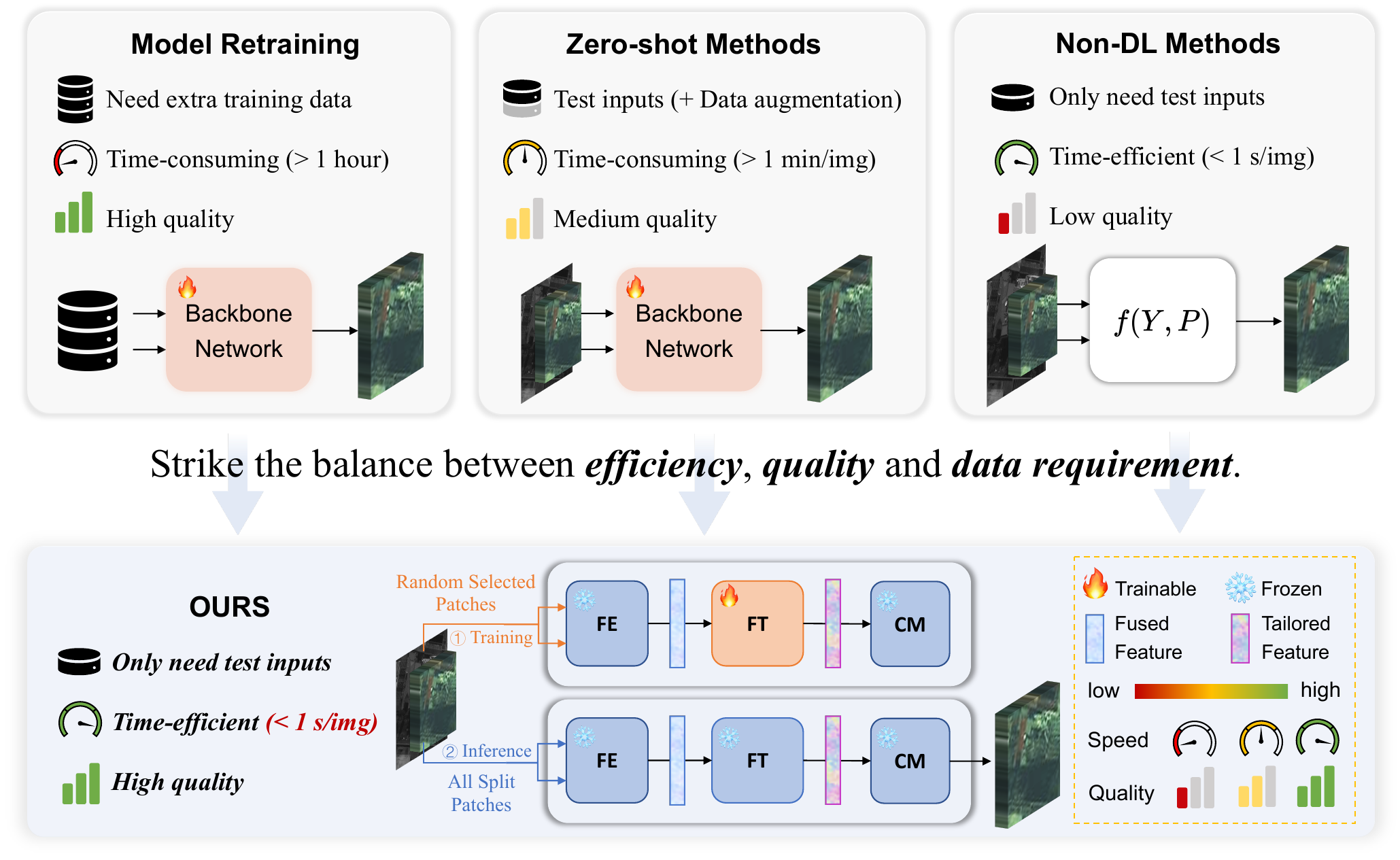}
    \caption{Comparison of different cross-sensor pansharpening methods. Existing approaches (upper panel) struggle to balance efficiency, quality, and data requirement. In contrast, our method (lower panel) achieves state-of-the-art performance with sub-second efficiency, requires only test-time inputs, and fully leverages pretrained model capabilities.}
  \label{fig:header_fig}
\end{figure}

Pansharpening methods are broadly classified into four categories: component substitution, multi-resolution analysis, variational optimization, and deep learning (DL) approaches \cite{caotpami2025, Cao_2025_ICCV}. The first three categories are traditional methods that process inputs via mathematical transformations without model training, but their inability to capture nonlinear PAN–HRMS relationships limits fusion quality \cite{xiao2022, xiao2023variational}. In contrast, deep learning methods \cite{10003241, liu2024promptfusion} leverage excellent feature extraction capability of neural networks, such as convolutional neural network (CNN)-based models \cite{dicnn, cao2021pancsc, wang2024deep}, Transformer-based models \cite{fusformer, zhang2021gtp}, and Diffusion-based methods \cite{cao2024diffusion, zhong2024ssdiff}.

While DL methods have achieved impressive results, the growing diversity of satellite sensors makes \textit{cross-sensor generalization increasingly important} for existing pansharpening models. However, \textit{current DL methods often suffer from cross-sensor degradation}, where models pretrained on data from a specific sensor generalize poorly to data from others. As shown in Figure \ref{fig:header_fig} (upper panel), typical strategies addressing the issue include model retraining or zero-shot methods, but they are highly time-consuming or even need extra training data. Zero-shot methods \cite{psdip, zspan} usually take minutes to process only one input pair ($512\times512$ PAN), let alone model retraining that takes hours. And although traditional methods are more efficient, their fusion quality remains significantly lower than that of DL methods. More importantly, most existing methods often disrupt the pretrained model weights, failing to fully leverage the rich representational capacity already embedded in them.
\textit{In sum, current methods for addressing cross-sensor degradation struggle to balance the trade-off between quality, efficiency, and data requirement, while also underutilizing the capabilities of existing pretrained models.}

Therefore, two critical challenges in tackling cross-sensor degradation for real-world pansharpening remain underexplored: 1) \textit{minimizing generalization cost by reducing processing time and avoiding the need for extra training data}; 2) \textit{preserving the pretrained model’s existing capabilities while ensuring improved generalization to unseen sensors}. 
To address the challenges, we propose a plug-and-play solution to tackle cross-sensor degradation with sub-second efficiency: the \textbf{Efficient Residual Feature Tailoring (ERFT)} pipeline. 
It begins by modularly decomposing pansharpening models into two general components: Feature Extractor (FE) and Channel Mapper (CM). 
The FE generates high-dimensional fused feature from the LRMS-PAN input with various neural network architectures. And the CM maps the feature from feature space to channel space to generate the high resolution output image, mostly with a shallow CNN structure. 
Then we integrates a \textbf{Feature Tailor (FT)} between frozen FE and CM to address cross-sensor degradation of pretrained model at feature level while leveraging capabilities of pretrained models. 
To maximize efficiency and ensure structural compatibility, the FT also employs a CNN structure, which is trained with physics-aware unsupervised losses to preserve spectral and spatial fidelity. 
To further improve runtime efficiency, our method operates in a patch-wise manner during both training and inference. Specifically, the LRMS-PAN input is first partitioned into multiple patches. A randomly selected subset is used to train the FT, while keeping the pretrained FE and CM frozen. Then all modules (FE, FT, and CM) are fixed to perform parallel inference on all patches. Finally, the predicted HRMS patches are stitched together to form the final high-resolution output.

Experiments on real-world data from sensors with different band setting again prove that our method tackles cross-sensor degradation with \textit{state-of-the-art (SOTA) quality and efficiency}. Moreover, our method could process input images scaled up to megapixel level (images with over one million pixels). In detail, with a commonly used RTX 3090 GPU, our method can complete pansharpening of $512\times512\times8$ image within \textit{0.2 seconds} and $4000\times4000\times8$ within \textit{3 seconds} at the fastest setting, which is \textit{over 100 times faster} than zero-shot methods and outperform latest pansharpening results of 2025. To the best of our knowledge, our method is the \textit{first} to tackle cross-sensor degradation with the \textit{sub-second efficiency} while achieving SOTA fusion quality. 

The main contributions are summarized as follows:

\begin{itemize}
  \item We propose the {Efficient Residual Feature Tailoring (ERFT)} to address the efficiency challenge of cross-sensor generalization. It avoids the need for extra training data and achieves training and inference within a sub-second runtime that has not been attained in this domain.

  \item Our method offers a novel plug-and-play solution for cross-sensor degradation in real-world pansharpening. It avoids modifying the pretrained model parameters, thereby preserving their existing capabilities while enabling improved generalization to unseen sensors.

  \item Extensive experiments demonstrate that our method achieves state-of-the-art results on both in-sensor and cross-sensor datasets with sub-second efficiency. Its ability to process megapixel-scale inputs within 3 seconds further showcases its scalability and practical potential.
\end{itemize}

\section{Related Works}
\subsection{Deep Learning-based Pansharpening}

Deep learning (DL)-based methods have significantly advanced the task of pansharpening by leveraging diverse neural network architectures to learn complex nonlinear mappings and extract rich spatial-spectral representations. CNN-based models such as PanNet \cite{pannet}, DiCNN \cite{dicnn}, and FusionNet \cite{fusionnet} utilize convolutional layers to capture hierarchical features, and have demonstrated strong performance in enhancing spatial detail while maintaining spectral fidelity. Building on these foundations, Transformer-based models \cite{panformer, fusformer} and diffusion-based approaches \cite{pandiff, ssdiff} introduce more expressive architectures that can better model long-range dependencies and complex input distributions. These models enable deeper and more flexible nonlinear transformations, leading to more accurate and robust spatial-spectral fusion. \textit{Despite their impressive results, most deep learning-based methods suffer from evident cross-sensor degradation.} Specifically, models pretrained on data from a specific sensor often generalize poorly to data from other sensors due to data distribution shifts in spectral and spatial characteristics. This significantly limits their practical applicability in real-world scenarios involving diverse satellite sources.

\subsection{Cross-Sensor Application of Pansharpening}

Applying pansharpening models across different satellite sensors presents significant challenges due to variations in spectral responses and spatial characteristics. To address this, existing methods typically follow three main strategies:

(1) \textbf{Model Retraining}: DL models are retrained on data from the target sensor. While effective, this approach requires substantial training time (often several hours) and access to extra training data on the new sensor, and often fails to fully leverage the pretrained model’s existing knowledge.

\begin{figure*}[ht]
  \centering
  \includegraphics[width=0.85\textwidth]{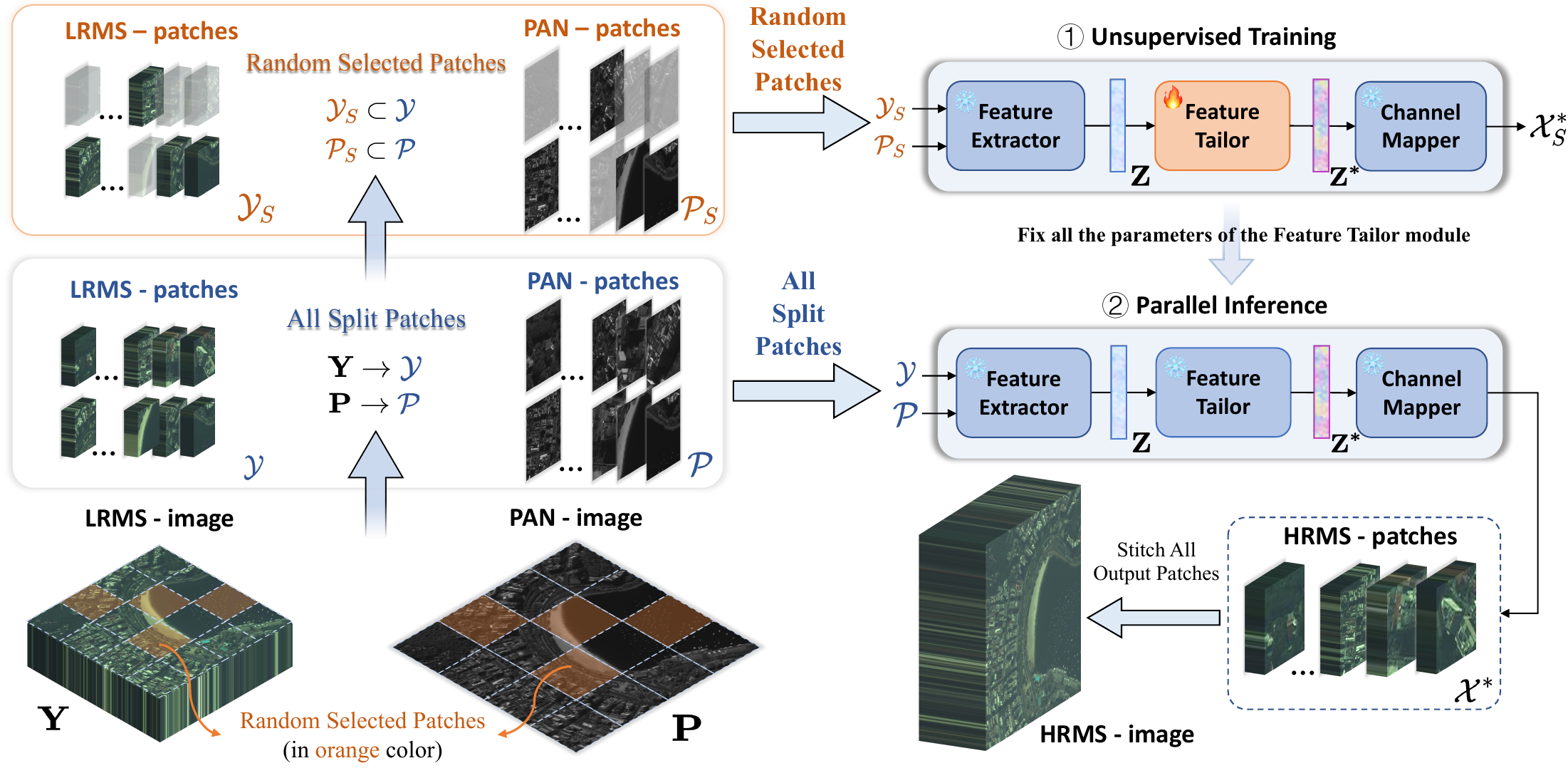}
  \caption{Our Efficient Residual Feature Tailoring pipeline is conducted in a patch-wise manner. Specifically, (1) Random selected patches are selected for unsupervised training of the Feature Tailor, enabling the feature-level adjustments; (2) Parallel inference is conducted on all split patches, whose resulting HRMS patches are stitched together to form the final HRMS image. }
  \label{overall_framework}
\end{figure*}

(2) \textbf{Zero-Shot Methods}: These methods \cite{psdip, zspan} optimize pansharpening models during inference without requiring supervision. They eliminate the need for extra training data and maintain relatively high fusion quality. However, they typically require several minutes to process a single LRMS-PAN input, highlighting the efficiency challenges of cross-sensor generalization.

(3) \textbf{Traditional Methods}: Traditional non-DL approaches such as Component Substitution (CS) \cite{cs1, cs2}, Multiresolution Analysis (MRA) \cite{mra1, mra2}, and Variational Optimization (VO) \cite{vo1, vo2} are efficient and require only a single PAN-LRMS pair as the input, but their fusion quality lags significantly behind modern deep learning methods.

\textit{In summary, existing strategies struggle to balance the trade-off between quality, efficiency, and data requirements, highlighting the challenge of efficient cross-sensor generalization without requiring additional training data.}

\section{Methodology}

As shown in Figure \ref{overall_framework}, we propose the plug-and-play pipeline named \textbf{Efficient Residual Feature Tailoring (ERFT)} to efficiently address cross-sensor degradation in pansharpening. Our method comprises three components: (1) residual feature tailoring for cross-sensor generalization; (2) patch-wise training and inference for improved efficiency; (3) physics-aware unsupervised losses for spatial and spectral fidelity.

\subsection{Residual Feature Tailoring} \label{sec:position_and_structure}

In this section, we detail the design of our \textbf{Feature Tailor (FT)} module, including both its placement and structure. 

\subsubsection{(I) Tailoring Position Design via Modular Decomposition}

DL-based pansharpening models generally take an LRMS image $\mathbf{Y}$ and a PAN image $\mathbf{P}$ as inputs, and produces a $C$-channel HRMS output $\hat{\mathbf{X}}$, which can be described as:
\begin{equation}
\hat{\mathbf{X}} = \mathcal{F}(\mathbf{Y}, \mathbf{P}; \theta), \quad \mathbf{\hat X} \in \mathbb{R}^{C \times H \times W},
\label{ori_infer}
\end{equation}
where $\mathcal{F}$ denotes the backbone network and $\theta$ its parameters.

Despite architectural variations, most existing methods \cite{pannet, dicnn, pmac} share a common modular pattern: an early-stage feature extractor $\mathcal{F}_1$ that encodes and fuses input information, and a late-stage channel mapper $\mathcal{F}_2$ that projects the fused features back into the output space. This decomposition yields:
\begin{equation}
\mathcal{F} = \mathcal{F}_2 \circ \mathcal{F}_1, \quad \theta = \theta_1 \cup \theta_2,
\end{equation}
where $\theta_1$, $\theta_2$ are the parameters of $\mathcal{F}_1$ and $\mathcal{F}_2$, respectively.

The feature extractor $\mathcal{F}_1$ transforms inputs into a high-dimensional representation $\mathbf{Z}$ with $S$ latent dimensions:
\begin{equation}
\mathbf{Z} = \mathcal{F}_1(\mathbf{Y}, \mathbf{P}; \theta_1), \quad \mathbf{Z} \in \mathbb{R}^{S \times H \times W}.
\label{infer1}
\end{equation}

The channel mapping stage then reconstructs the HRMS image $\hat{\mathbf{X}}$ from $\mathbf{Z}$, typically with a residual shortcut from the upsampled LRMS image $\mathbf Y$ to maintain spectral fidelity:
\begin{equation}
\hat{\mathbf{X}} = \mathcal{F}_2(\mathbf{Z}; \theta_2) + \texttt{UpSample}(\mathbf{Y}).
\label{channel_transformation}
\end{equation}

\begin{figure}[ht]
    \centering
    \includegraphics[width=0.9\linewidth]{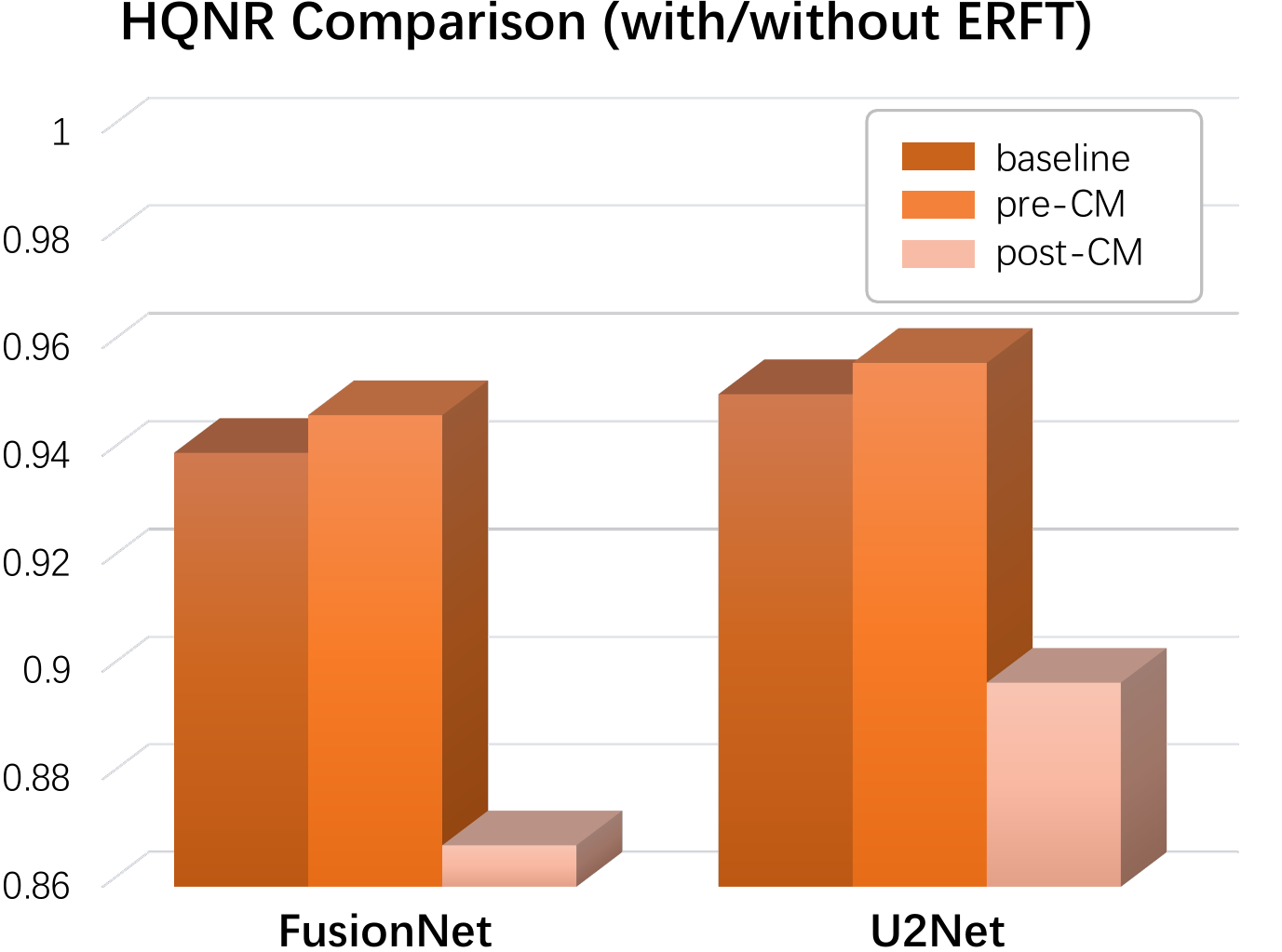}
    \caption{HQNR comparison for FusionNet and U2Net on WV3 dataset under three FT placement strategies: FT not inserted (Baseline), FT inserted before channel mapping (pre-CM), and FT inserted after channel mapping (post-CM). The pre-CM configuration consistently outperforms the others, validating the effectiveness of feature-level adjustments.}
    \label{fig:position_cmp}
\end{figure}

We insert the \textbf{Feature Tailor (FT)} module at the junction between $\mathcal{F}_1$ and $\mathcal{F}_2$.  This critical interface carries the fused feature $\mathbf Z$, which is rich in information from both modalities yet unconstrained by the output space, thus serving as a favorable target for cross-sensor generalization. This choice is also supported by prior studies \cite{dou2019domain_generalization, yosinski2014feature_transferbility}, which show that intermediate features possess greater transferability and representational flexibility than shallow or output-level features. Moreover, our experiments further validate this design: as shown in Figure~\ref{fig:position_cmp}, feature-space adjustments by placing the FT before the channel mapper (pre-CM) consistently outperform output-space adjustments by placing the FT after the channel mapper (post-CM) and the baseline, highlighting this interface as an ideal and effective position for cross-sensor generalization.

\subsubsection{(II) Structural Design of the Feature Tailor} 

The Feature Tailor (FT) module is implemented as a residual block with shallow CNN. Positioned at the intermediate interface between the feature extractor $\mathcal{F}_1$ and the channel mapper$ \mathcal{F}_2$, it operates on the fused latent representation$ \mathbf{Z}$, which encodes rich spatial and spectral cues from both input modalities.

Instead of generating new representations from scratch, the FT module learns a residual adjustment $\Delta(\mathbf{Z})$ using unsupervised losses to adapt $\mathbf{Z}$ to the test-time distribution. Formally, the tailored feature  $\mathbf{Z}^*$ is computed as:
\begin{equation}
\mathbf{Z}^* = \mathbf{Z} + \mathcal{G}(\mathbf{Z}; \phi), \quad \mathbf{Z}^* \in \mathbb{R}^{S \times H \times W}, 
\label{Z_tailoring}
\end{equation}
where $\mathcal{G}(\cdot)$ is a shallow CNN network with parameters $\phi$.

The tailored feature $\mathbf{Z}^*$ is then passed to the frozen channel mapper $\mathcal{F}_2$ to reconstruct the final HRMS output:
\begin{equation}
\hat{\mathbf{X}}^* = \mathcal{F}_2(\mathbf{Z}^*; \theta_2) + \texttt{UpSample}(\mathbf{Y}), 
\end{equation}
where $\hat{\mathbf{X}}^* \in \mathbb{R}^{C \times H \times W}$ denotes the ERFT-enhanced output.

This residual design is motivated by three key considerations. First, it allows the FT module to make precise and efficient adjustments to the latent features without disrupting the pretrained backbone. Second, using a shallow CNN ensures both fast optimization and structural continuity, as the following channel mapper $\mathcal{F}_2$ is also typically implemented as a shallow CNN. Finally, the design aligns with findings from existing works \cite{ilyas2019bug, vonkugelgen2025purturbation}, which demonstrate that even small perturbations to intermediate features can significantly alter predictions, suggesting that learning residual feature corrections is an effective means of improving cross-sensor generalization.

\subsection{Efficient Patch-wise Training and Inference}

To improve runtime efficiency, we introduce a patch-wise strategy for both training and inference. 
As shown in Algorithm \ref{algo_workflow}, rather than processing the entire image at once, each PAN–LRMS input pair is partitioned into multiple patches. A randomly selected subset of these patches is then used to train the FT module in an unsupervised manner. Once the training is completed, the FT parameters are fixed and subsequently applied to inference on all split patches. For clarity, we omit the outer loop over the training epochs for Lines 3 to 11 in Algorithm~\ref{algo_workflow}, though it is used in practice.
\begin{algorithm}
\caption{Patch-wise Workflow of ERFT}
\label{algo_workflow}
\begin{algorithmic}[1]
\Statex \textbf{Input: }PAN image $\mathbf{P}$, LRMS image $\mathbf{Y}$, backbone net $\mathcal F$
\Statex \textbf{Output: }HRMS output image $\hat{\mathbf{X}}^*$
\State Split $\mathbf{Y}, \mathbf{P}$ into $N$ patches: $\{(\mathbf{Y}_i, \mathbf{P}_i)\}_{i=1}^N$ 
\State Randomly select $M$ training patches: $\mathcal{T} \subset \{1,\dots,N\}$ 
\State Initialize loss: $\mathcal{L} \gets 0$
\For{each $i \in \mathcal{T}$} \Comment{Training on selected patches}
    \State $\mathbf{Z}_i \gets \mathcal{F}_1(\mathbf{Y}_i, \mathbf{P}_i)$ 
    \State $\mathbf{Z}^*_i \gets \mathcal{G}(\mathbf{Z}_i) + \mathbf{Z}_i$ \Comment{Residual feature tailoring}
    \State $\hat{\mathbf{X}}_i \gets \mathcal{F}_2(\mathbf{Z}^*_i) + \mathrm{UpSample}(\mathbf{Y}_i)$
    \State $\hat{\mathbf{X}}_i^0 \gets \mathcal{F}_2(\mathbf{Z}_i) + \mathrm{UpSample}(\mathbf{Y}_i)$
    \State $\mathcal{L} \gets \mathcal{L} + \mathcal{L}_{\text{unsup}}(\hat{\mathbf{X}}_i, \hat{\mathbf{X}}_i^0, \mathbf{Y}_i, \mathbf{P}_i)$
\EndFor
\State Update $\mathcal{G}$ using accumulated loss $\mathcal{L}$
\State Freeze $\mathcal{G}$
\For{each $i = 1$ to $N$} \Comment{Inference on all patches}
    \State $\mathbf{Z}_i \gets \mathcal{F}_1(\mathbf{Y}_i, \mathbf{P}_i)$ 
    \State $\mathbf{Z}^*_i \gets \mathcal{G}(\mathbf{Z}_i) + \mathbf{Z}_i$ \Comment{Residual feature tailoring}
    \State $\hat{\mathbf{X}}_i \gets \mathcal{F}_2(\mathbf{Z}^*_i) + \mathrm{UpSample}(\mathbf{Y}_i)$
\EndFor
\State Get the final output HRMS image $\hat{\mathbf{X}}^* \gets \mathrm{Stitch}(\{\hat{\mathbf{X}}_i\})$
\end{algorithmic}
\end{algorithm}

\begin{table}[h]
\centering
\begin{tabular}{l|c|c|c}
\toprule
\textbf{Architecture} & \textbf{Before} & \textbf{After} & \textbf{Speedup} \\
\midrule
CNN        & \(\mathcal{O}(H W)\)             & \(\mathcal{O}(\frac{M}{B} h w)\)             & \(\frac{N}{M} B\) \\
Attention & \(\mathcal{O}(H^2 W^2)\)             & \(\mathcal{O}(\frac{M}{B} h^2 w^2)\)             & \(\frac{N^2}{M}B\) \\
\midrule
CNN        & \(\mathcal{O}(H W)\)             & \(\mathcal{O}(\frac{N}{B} h w)\)   & \(B\) \\
Attention& \(\mathcal{O}(H^2 W^2)\)             & \(\mathcal{O}(\frac{N}{B} h^2 w^2)\)   & \(N \cdot B\) \\
\bottomrule
\end{tabular}
\caption{
Theoretical complexity and speedup of our patch-wise strategy for CNN and Attention architectures. 
“Before” denotes complexity of full-image processing; “After” denotes complexity of patch-wise processing. 
For training (upper block), only \(M\) out of \(N\) patches are used; for inference (lower block), all \(N\) patches are processed in parallel. 
Here, \(H \times W\) is the input image size, \(h \times w\) is the patch size.
}
\label{tab:patchwise_comparison}
\end{table}

\begin{figure*}[ht]
  \centering
  \includegraphics[width=0.8\textwidth]{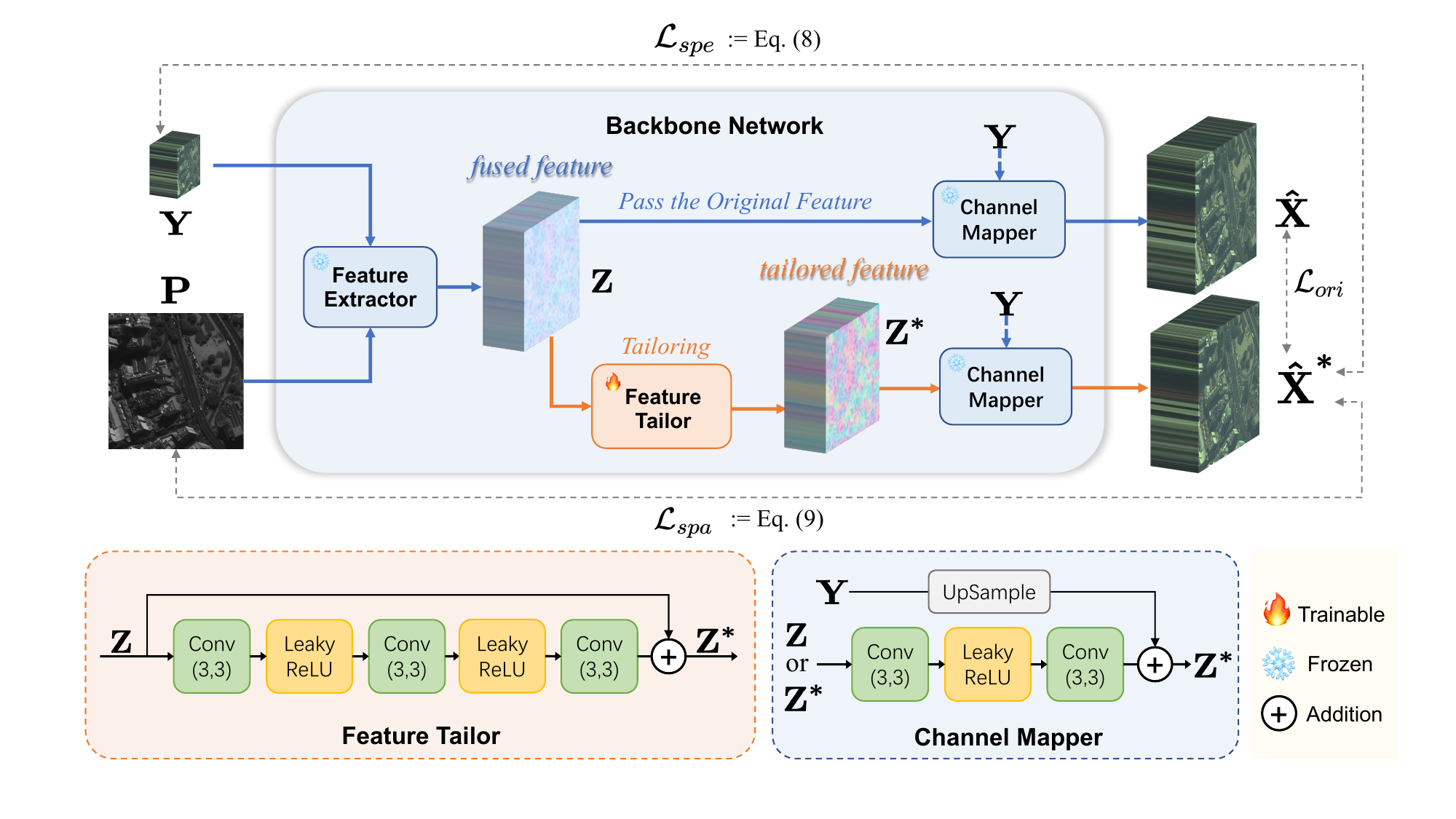}
  \caption{Detailed workflow of unsupervised training. The LRMS image $\mathbf{Y}$ and PAN image $\mathbf{P}$ are fed into the backbone network to extract high-dimensional latent features $\mathbf{Z}$, which are then refined by the FT module to produce tailored features $\mathbf{Z}^*$. Both $\mathbf{Z}$ and $\mathbf{Z}^*$ are passed through the channel mapping (CM) module to generate the original and tailored HRMS outputs, $\mathbf{\hat{X}}$ and $\mathbf{\hat{X}}^*$, respectively. These outputs are compared with the inputs to compute unsupervised losses to update the FT module.}
  \label{fig:network_details}
\end{figure*}

Such patch-wise strategy for both training and inference greatly reduces the time complexity. By training the FT module on only a small subset of patches, it lowers the number of forward and backward passes, accelerating optimization. Moreover, since patches are processed independently, they can be grouped into batches and executed in parallel during both training and inference, further improving efficiency. This strategy is especially beneficial for backbone networks with superlinear time complexity, such as Transformers. The theoretical complexities and speedup gains of our patch-wise strategy for typical CNN-based and attention-based architectures are reported in Table \ref{tab:patchwise_comparison}, with formal derivations provided in the \textit{extended version}.

\subsection{Unsupervised FT Training Loss} \label{sec:updating_method}
To address cross-sensor degradation using only test inputs, we optimize our framework (Fig. \ref{fig:network_details}) with the unsupervised loss from Eq.\eqref{equ:total_loss}, which combines spectral ($\mathcal{L}_{spe}$), spatial ($\mathcal{L}_{spa}$), and original-output consistency ($\mathcal{L}_{ori}$) losses. Patch indices are omitted from the following formulas for clarity.
\begin{equation}
  \mathcal{L}_{total} = \eta_{1}\mathcal{L}_{spe} + \eta_{2}\mathcal{L}_{spa} + \eta_{3}\mathcal{L}_{ori}, 
\label{equ:total_loss}
\end{equation}
where $\eta_1$, $\eta_2$, and $\eta_3$ balance the contribution of each term.

\subsubsection{(I) Spectral Loss ($\mathcal L_{spe}$)}
This term preserves spectral fidelity by encouraging the blurred and downsampled HRMS output $\mathbf{\hat X^*}$ to match the LRMS input $\mathbf{Y}$, where $\mathbf{B}$ denotes the modulation transfer function (MTF)-based blur kernel:
\begin{equation}
  \label{spectral_loss}
  \mathcal{L}_{spe} = \|\mathrm{DownSample}(\mathbf{\hat X^*} \mathbf B) - \mathbf{Y}\|_1.
\end{equation}

\subsubsection{(II) Spatial Loss ($\mathcal{L}_{spa}$)}
This term promotes spatial consistency by aligning the high-frequency details of $\mathbf{\hat X^*}$ and PAN $\mathbf{P}$, following method in \cite{vo+net}. In Eq. \eqref{spatial_loss}, $\mathbf{\hat P}$ denotes the PAN broadcasted to $C$ channels, $\circ$ is element-wise multiplication, and $\oslash$ is element-wise division:
\begin{equation}
  \label{spatial_loss}
  \mathcal{L}_{spa} = \|\mathbf{\hat X^*} - \mathbf{\hat X^*} \mathbf B \circ (\mathbf{\hat P} \oslash \mathbf{\hat P} \mathbf B)\|_1.
\end{equation}

\subsubsection{(III) Consistency Loss ($\mathcal{L}_{ori}$)}
This term prevents overfitting by regularizing the ERFT-enhanced output to remain close to the original prediction from the frozen backbone:
\begin{equation}
  \label{original_loss}
  \mathcal{L}_{ori} = \|\mathbf{\hat X^*} - \mathbf{\hat X}\|_1.
\end{equation}

\begin{table*}[htbp]
  \centering 
  \small
  \setlength{\tabcolsep}{2.7pt}
  \renewcommand\arraystretch{1.2}
    \begin{tabular}{c|ccc|ccc}
      \toprule
      \toprule
      \multirow{2.2}{*}{Method} & \multicolumn{3}{c|}{QB $\to$ GF-2 (4-band sensor): Avg$\pm$std} & \multicolumn{3}{c}{WV-3 $\to$ WV-2 (8-band sensor): Avg$\pm$std} \\
      \cmidrule{2-4}\cmidrule{5-7}
       & HQNR$\uparrow$ & $D_\lambda\downarrow$ & $D_s\downarrow$ & HQNR$\uparrow$ & $D_\lambda\downarrow$ & $D_s\downarrow$ \\
      \midrule
          BT-H         & 0.7293$\pm$0.0253 & 0.1559$\pm$0.0239 & 0.1359$\pm$0.0192 & 0.8300$\pm$0.0430 & 0.0860$\pm$0.0301 & 0.0925$\pm$0.0208 \\
          C-BDSD      & 0.6996$\pm$0.0323 & 0.2149$\pm$0.0302 & 0.1091$\pm$0.0154 & 0.6956$\pm$0.0461 & 0.2253$\pm$0.0488 & 0.1019$\pm$0.0296 \\
          BDSD-PC      & 0.6762$\pm$0.0378 & 0.1971$\pm$0.0346 & 0.1578$\pm$0.0279 & 0.8286$\pm$0.0432 & 0.1413$\pm$0.0320 & $\second{0.0356}$$\pm$$\second{0.0213}$ \\
          MTF-GLP       & 0.7151$\pm$0.0366 & 0.1592$\pm$0.0254 & 0.1495$\pm$0.0355 & 0.8549$\pm$0.0475 & 0.0582$\pm$0.0221 & 0.0930$\pm$0.0320 \\
          MTF-GLP-FS  & 0.7423$\pm$0.0338 & 0.1438$\pm$0.0251 & 0.1331$\pm$0.0291 & 0.8658$\pm$0.0415 & 0.0563$\pm$0.0212 & 0.0830$\pm$0.0260 \\
          MF                 & 0.7817$\pm$0.0231 & \best{0.0600}$\pm$\best{0.0260} & 0.1470$\pm$0.0223 & 0.8508$\pm$0.0538 & 0.0704$\pm$0.0308 & 0.0857$\pm$0.0297 \\
          PsDip          & 0.7825$\pm$0.0325 & 0.1323$\pm$0.0286 & 0.0981$\pm$0.0256 & 0.8980$\pm$0.0226 & \best{0.0385}$\pm$\best{0.0239} & 0.0659$\pm$0.0158 \\
          ZS-Pan        & $\second{0.8925}$$\pm$$\second{0.0240}$ & 0.0778$\pm$0.0182 & \best{0.0323}$\pm$\best{0.0159} & $\second{0.9112}$$\pm$$\second{0.0336}$ & 0.0476$\pm$0.0270 & 0.0435$\pm$0.0130 \\
          FusionMamba  & 0.7627$\pm$0.0724 & 0.1627$\pm$0.0695 & 0.0893$\pm$0.0352 & 0.9022$\pm$0.0242 & 0.0572$\pm$0.0272 & 0.0429$\pm$0.0112 \\
          WFANet         & 0.7093$\pm$0.0450 & 0.1975$\pm$0.0497 & 0.1154$\pm$0.0382 & \best{0.9128}$\pm$\best{0.0301} & 0.0526$\pm$0.0302 & 0.0366$\pm$0.0054 \\
          FusionNet   &  0.8397$\pm$0.0509 & 0.1114$\pm$0.0500 & 0.0551$\pm$0.0137 & 0.8881$\pm$0.0213 & 0.0543$\pm$0.0273 & 0.0606$\pm$0.0162 \\
          U2Net          & 0.7089$\pm$0.0535 & 0.2209$\pm$0.0574 & 0.0909$\pm$0.0286 & 0.8706$\pm$0.0809 & 0.0936$\pm$0.0790 & 0.0400$\pm$0.0098 \\
          \midrule
          ERFT$_{\text{FusionNet}}$     & \best{0.8970}$\pm$\best{0.0471} & $\second{0.0706}$$\pm$$\second{0.0382}$ & $\second{0.0353}$$\pm$$\second{0.0145}$     &    
          0.9100$\pm$0.0255 & $\second{0.0445}$$\pm$$\second{0.0214}$ & 0.0475$\pm$0.0160 \\
          ERFT$_{\text{U2Net}}$        & 0.8264$\pm$0.1175 & 0.0930$\pm$0.1279 & 0.0888$\pm$0.0021 & \best{0.9128}$\pm$\best{0.0526} & 0.0617$\pm$0.0494 & \best{0.0274}$\pm$\best{0.0073} \\
      \bottomrule
      \bottomrule
    \end{tabular}
  \caption{Performance comparison in cross-sensor scenarios on real-world datasets. “A $\to$ B” indicates that the model is pretrained on dataset A and tested on dataset B. All results are averaged over 20 test images. (\textbf{Bold}: best; \underline{Underline}: second best)}
  \label{tab:main_results}
\end{table*}

\section{Experiments}
 
\subsection{Experiment Settings}

\subsubsection{(I) Datasets and Metrics}
We conduct experiments on four datasets derived from the PanCollection \footnote{\url{https://github.com/liangjiandeng/PanCollection}}. These datasets are captured by WorldView-3 (WV-3), WorldView-2 (WV-2), QuickBird (QB), and Gaofen-2 (GF2), and consist of paired PAN and LRMS images, with each PAN image sized at $512 \times 512$ pixels.
Our method is specifically designed for real-world pansharpening, which is of paramount practical significance. Accordingly, we primarily evaluate performance on real-world data using three widely adopted reference-free metrics: HQNR \cite{aiazzi2014full}, $D_s$, and $D_{\lambda}$. Notably, HQNR is derived from $D_s$ and $D_{\lambda}$, providing an assessment of both spatial and spectral fidelity.

\subsubsection{(II) Benchmarks}
We compare our method with several SOTA pansharpening approaches, including \textit{six traditional methods}, BT-H \cite{BT-H}, C-BDSD \cite{C-BDSD}, BDSD-PC \cite{BDSD-PC}, MTF-GLP \cite{MTF-GLP}, MTF-GLP-FS \cite{MTF-GLP-FS}, and MF \cite{MF}; and \textit{six DL-based methods}, FusionNet \cite{fusionnet}, U2Net \cite{u2net}, PsDip \cite{psdip}, ZS-Pan \cite{zspan}, FusionMamba \cite{fusionmamba}, and WFANet \cite{WFANet}. Then we employ FusionNet \cite{fusionnet} and U2Net \cite{u2net} as our baseline networks, representing typical CNN and attention-based architectures, respectively.

\subsubsection{(III) Implementation Details}
All experiments were conducted on a hardware setup comprising an NVIDIA RTX 3090 GPU with 24GB memory and Intel i9-12900 CPU.

\subsection{Comparison with State-of-the-Art Methods}
\begin{figure*}
    \centering
    \includegraphics[width=1\linewidth]{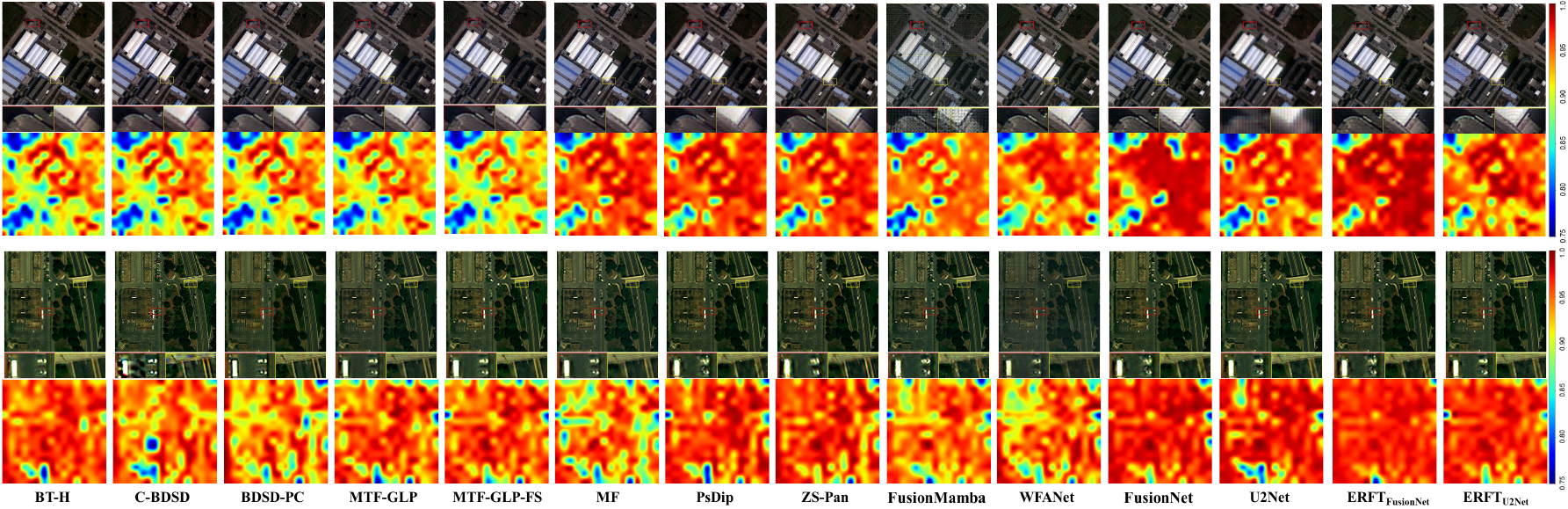}
    \caption{Visual Fusion Examples and HQNR Map in two cross-sensor cases: QB $\to$ GF2 (upper) and WV3 $\to$ WV2 (lower).}
    \label{fig:visual}
\end{figure*}

Our method specifically targets cross-sensor degradation in real-world pansharpening. Therefore we conduct evaluations in two cross-sensor cases: QB $\to$ GF2 and WV3 $\to$ WV2, which denotes models pretrained\footnote{Traditional and zero-shot methods do not require pretraining and can be directly applied in cross-sensor settings.} on data from the prior sensor and tested on data from the latter sensor. Quantitative results are reported in Table~\ref{tab:main_results}, while visual examples and corresponding HQNR maps are shown in Figure \ref{fig:visual}.

The results show that most DL-based methods, excluding zero-shot ones, struggle to generalize across sensors. They sometimes even underperform traditional methods, especially in the QB $\to$ GF2 case. In contrast, our method significantly improves the cross-sensor generalization of backbone models, setting a new SOTA in cross-sensor scenarios. For instance, HQNR improves by 5.73\% for FusionNet and 11.75\% for U2Net in the QB $\to$ GF2 case, surpassing zero-shot methods which are typically the most generalizable.

\subsection{Breaking the Runtime Efficiency Bottleneck}
As shown in Table \ref{tab:main_results}, zero-shot methods achieve the top-tier fusion quality in cross-sensor cases, establishing themselves as the leading generalization methods aside from our proposed method. However, their efficiency remains a significant bottleneck, often requiring several minutes to optimize for a single test input. In contrast, our method offers a much more favorable efficiency with better fusion quality. As reported in Table \ref{tab:time}, it not only outperforms zero-shot methods in fusion quality across most cases, but also reduces processing time to the sub-second level, which is over 100 times faster than zero-shot methods, and makes it substantially more practical for real-world applications.

\begin{table}[!ht]
    \centering
    \setlength{\tabcolsep}{6pt}
    \renewcommand\arraystretch{1.2}
    \resizebox{\linewidth}{!}{
    \begin{tabular}{c|l|cccc}
         \toprule
         Case & Method & HQNR$\uparrow$ & \(D_\lambda\downarrow\) & \(D_s\downarrow\) & Duration$\downarrow$ (s) \\
         \midrule
         
         \multirow{4}{*}{\shortstack{QB\\$\downarrow $\\GF-2}}  
         & PsDip             & 0.7825 & 0.1323 & 0.0981 & 282.59    \\
         & ZS-Pan            & \second{0.8925} & \second{0.0778} & \best{0.0323} & 66.74    \\
         & $\text{ERFT}_\text{FusionNet}$   & \best{0.8970} & \best{0.0706} & \second{0.0353} & \best{0.11}  \\
         & $\text{ERFT}_\text{U2Net}$       & 0.8264 & 0.0930 & 0.0888 & \second{0.77}  \\
         \midrule
         
         \multirow{4}{*}{\shortstack{WV-3\\$\downarrow $\\WV-2}}
         & PsDip           & 0.8980 & 0.0385 & 0.0659 & 276.18    \\
         & ZS-Pan            & \second{0.9112} & \second{0.0476} & \second{0.0435} & 67.50    \\
         & $\text{ERFT}_\text{FusionNet}$   & 0.9100 & \best{0.0445} & 0.0475 & \best{0.13}  \\
         & $\text{ERFT}_\text{U2Net}$       & \best{0.9128} & 0.0617 & \best{0.0274} & \second{0.89}  \\
         
         \bottomrule
    \end{tabular}
    }
    \caption{Comparison of efficiency and fusion quality in two cross-sensor cases between our method and zero-shot methods that typically yield top performance. All values are averaged over test inputs. (\best{Bold}: best; \second{Underline}: second best)}
    \label{tab:time}
\end{table}

\subsection{Ablation Study}
\begin{table}[t]
\centering
\small
\setlength{\tabcolsep}{3pt}
\renewcommand{\arraystretch}{1.15}
\begin{tabular}{c|cc|ccc|c}
\toprule
\textbf{Backbone} & \textbf{FT} & \textbf{PW} & HQNR$\uparrow$ & $D_\lambda\downarrow$ & $D_s\downarrow$ & {Duration (s)}$\downarrow$ \\
\midrule
\multirow{4}{*}{FusionNet}
&              &              & 0.8397 & 0.1114 & 0.0551 &  0.15 \\
& \checkmark   &              & 0.9002 & 0.0698 & 0.0322 &  0.45 \\
&              & \checkmark   & 0.8395 & 0.1112 & 0.0557 &  0.04 \\
& \checkmark   & \checkmark   & 0.8970 & 0.0706 & 0.0353 &  0.13 \\
\midrule
\multirow{4}{*}{U2Net}
&              &              & 0.7089 & 0.2209 & 0.0909 &  0.60 \\
& \checkmark   &              & 0.8308 & 0.0873 & 0.0897 &  7.53 \\
&              & \checkmark   & 0.7088 & 0.2211 & 0.0907 &  0.38 \\
& \checkmark   & \checkmark   & 0.8264 & 0.0930 & 0.0888 &  0.89 \\
\bottomrule
\end{tabular}
\caption{Ablation study on the effect of Feature Tailor (FT) and Patch-wise strategy (PW) under WV3 $\to$ WV2 case.}
\label{tab:ablation_wv2}
\end{table}

We conduct a series of ablation experiments to evaluate the contribution of two key components in our method: the feature tailoring (FT) for improving fusion quality and patch-wise (PW) strategy for boosting generalization efficiency. 

\subsubsection{(I) Feature Tailoring (FT)}
The FT module is removed from the ERFT-enhanced network to validate its importance in improving the generalization quality. As shown in Table \ref{tab:ablation_wv2}, both FusionNet and U2Net experience substantial HQNR drops (5.75\% and 11.76\%) after removing the FT module.

\subsubsection{(II) Patch-wise Strategy (PW)}
We also evaluate the impact of the PW strategy by disabling it and allowing the ERFT-enhanced network to train and infer on entire test inputs. As reported in Table~\ref{tab:ablation_wv2}, this setting significantly prolongs processing time, rising to 7.53 seconds for U2Net, thereby failing to achieve the desired sub-second efficiency.

\subsubsection{(III) Overall}
\begin{figure}[h]
    \centering
    \includegraphics[width=0.7\linewidth]{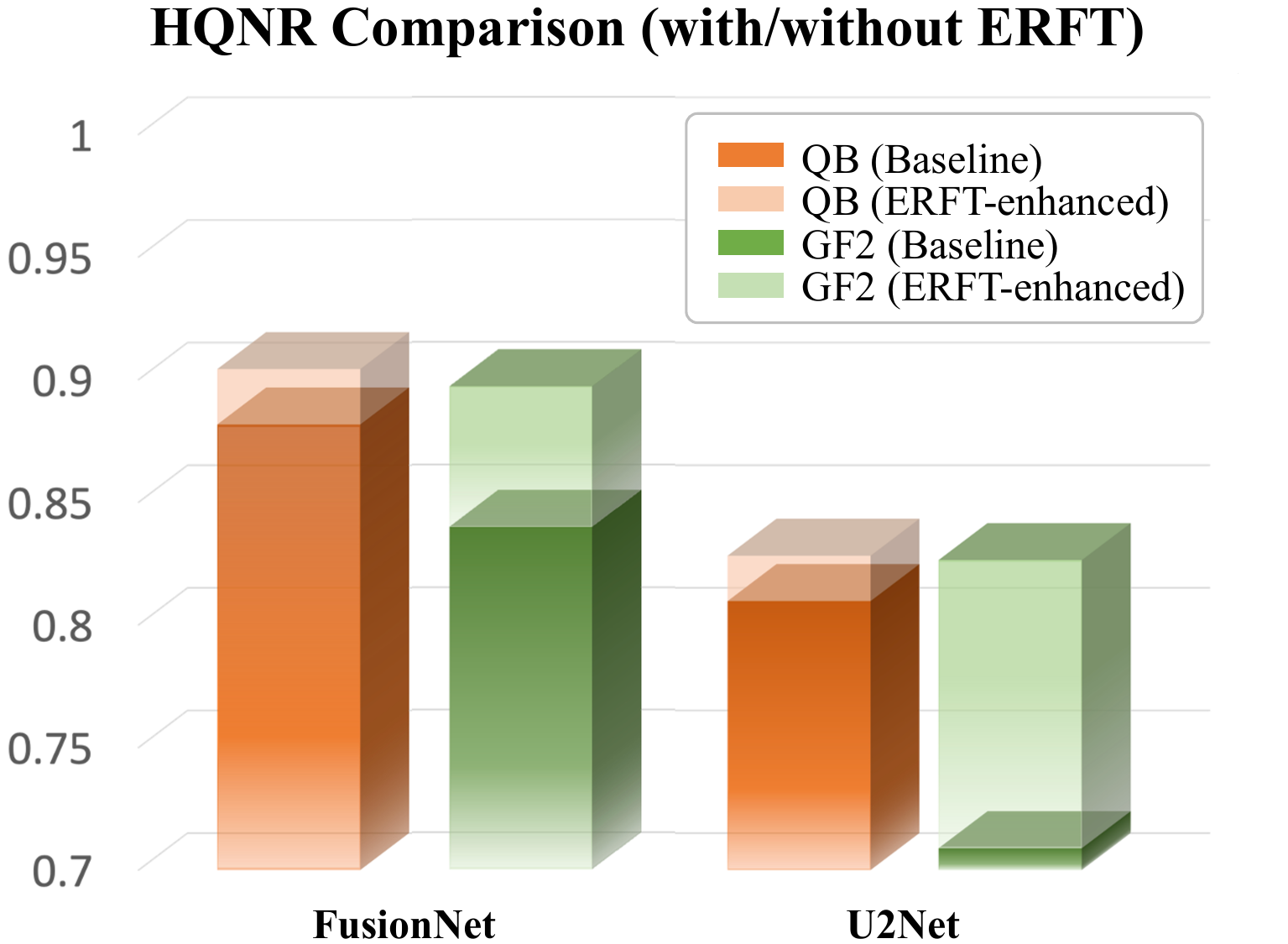}
    \caption{HQNR comparison of FusionNet and U2Net (both pretrained on QB) with and without ERFT enhancement on QB (in-sensor) and GF2 (cross-sensor). The translucent portions indicate the performance gain contributed by ERFT.}
    \label{fig:ablation}
\end{figure}

Finally, we compare the ERFT-enhanced networks with their baseline counterparts under both in-sensor and cross-sensor scenarios to evaluate their overall generalization performance. As illustrated in Figure \ref{fig:ablation}, the translucent portions of each bar indicate HQNR gains attributed to ERFT-enhancement. The results demonstrate consistent improvements in fusion quality across different scenarios. Moreover, the near-equal HQNR values between in-sensor and cross-sensor scenarios further highlight the strong generalization capability of our proposed method.


\subsection{Discussion and Extensive Application}
Apart from performance comparison and ablation, we also discuss the influence of hyperparameters including patch size, patch number, training epochs and the method's robostness on newer backbones like FusionMamba and WFANet. Moreover, we extend our method to the novel task of megapixel pansharpening, which processes input images exceeding one million pixels, where our method also achieves SOTA quality with second-level efficiency. Experimental settings and results are available in the \textit{extended version}.

\section{Conclusions}
This paper presents a novel approach to address cross-sensor degradation in real-world pansharpening: Efficient Residual Feature Tailoring (ERFT). Without relying on extra training data, ERFT achieves superior cross-sensor generalization within sub-second runtime, while fully leveraging pretrained backbones.
The method begins by identifying a critical interface via modular decomposition, at which the Feature Tailor module is inserted to learn residual corrections on the fused features, enabling efficient feature-level generalization to test inputs.
To further enhance efficiency, ERFT employs a patch-wise strategy during both training and inference, allowing batched parallel processing. 
This design not only accelerates computation significantly but is also theoretically justified.
Extensive experiments demonstrate that ERFT consistently outperforms existing methods in cross-sensor scenarios, achieving SOTA quality with over 100× speedup compared to leading zero-shot methods. Moreover, ERFT scales well to megapixel images while maintaining both quality and efficiency. These results highlight ERFT’s strong potential for practical deployment, offering a powerful solution for real-world cross-sensor pansharpening.

\section{Acknowledgments}
This research is supported by the Project of the Department of Science and Technology of Sichuan Province (Grant No. 2025YFNH0001).

\bibliography{aaai2026}

\clearpage
\appendix

\begin{abstract}
This document provides supplementary material for the main paper. We first present formal derivations of the theoretical time complexity and speedup achieved by our patch-wise strategy. Next, we offer additional implementation details, including key parameters and algorithmic settings. We further conduct extended ablation and discussion experiments to analyze the influence of various factors such as the number of patches, patch size, and training epochs, as well as evaluate performance under additional scenarios. Finally, we validate the scalability of our method, demonstrating that it can effectively perform pansharpening on inputs with a PAN size of $4000 \times 4000$, yielding HRMS outputs with 128 million pixels within 3 seconds in the fastest configuration, while maintaining superior fusion quality.
\end{abstract}

\section{Theoretical Time Complexity and Speedup}
To boost generalization efficiency, we propose a patch-wise strategy for both training and inference. Specifically, instead of directly processing an image of size $H \times W$, it is first partitioned into $N$ patches, with $M$ of them selected for training and all patches used during inference. To quantify the efficiency gains introduced by the proposed strategy, we analyze its theoretical time complexity and speedup under both CNN-based and attention-based architectures. 
For analytical clarity, we assume the average per-pixel processing time is the same between training and inference, and additional factors such as data loading, patch stitching, and model instantiation are not accounted for in the theoretical analysis.

\subsection{CNN-based Backbone}
For standard convolutional operations, each output pixel is computed by performing element-wise multiplications between the convolution kernel and a local receptive field in the input. This leads to a linear time complexity with respect to the number of pixels.

\subsubsection{(I) Full-image Processing}
Consider a convolutional layer with kernel size $K \times K$, input channels $C_{\text{in}}$, and output channels $C_{\text{out}}$. Let $t_1$ denote the average processing time per pixel in a CNN-based network. Then, the time required to process an entire image of size $H \times W$ is:
\begin{equation}
    T_{\text{full}} = H \cdot W \cdot K^2 \cdot C_{\text{in}} \cdot C_{\text{out}} \cdot t_1.
\end{equation}

Moreover, since $K$, $C_{\text{in}}$, and $C_{\text{out}}$ are treated as constants, the time complexity simplifies to:
\begin{equation}
    \mathcal{O}(HW).
\end{equation}

\subsubsection{(II) Patch-wise Training}
Since the input image is divided into $N$ patches of size $h \times w$, it follows that $H \cdot W = N \cdot h \cdot w$. During training, only $M$ out of the $N$ patches are selected. These $M$ patches can be grouped into batches of size $B$ for parallel processing, requiring approximately $\lceil \frac{M}{B} \rceil$ steps. Thus, the total time for patch-wise training is:
\begin{equation}
    T_{\text{patch}}^{\text{train}} = \left\lceil \frac{M}{B} \right\rceil \cdot h \cdot w \cdot K^2 \cdot C_{\text{in}} \cdot C_{\text{out}} \cdot t_1.
    \label{time_cnn_patch_train}
\end{equation}

Neglecting constants as before, the time complexity of CNN-based patch-wise training becomes:
\begin{equation}
    \mathcal{O}\left( \frac{M}{B} \cdot hw \right).
\end{equation}

The theoretical speedup of CNN-based patch-wise training over full-image training is then:
\begin{equation}
    S_{\text{train}}^{\text{CNN}} = \frac{T_{\text{full}}}{T_{\text{patch}}^{\text{train}}} \approx \frac{H \cdot W}{\frac{M}{B} \cdot h \cdot w} = \frac{N}{M} B.
\end{equation}

\subsubsection{(III) Patch-wise Inference}
Similarly, inference is performed on all $N$ patches. With batch size $B$, the inference time is:
\begin{equation}
    T_{\text{patch}}^{\text{infer}} = \frac{N}{B} \cdot h \cdot w \cdot K^2 \cdot C_{\text{in}} \cdot C_{\text{out}} \cdot t_2.
    \label{time_cnn_full_proc}
\end{equation}

Thus, the time complexity of CNN-based patch-wise inference is:
\begin{equation}
    \mathcal{O}\left( \frac{N}{B} \cdot hw \right).
\end{equation}

And the theoretical inference speedup over full-image inference becomes:
\begin{equation}
    S_{\text{infer}}^{\text{CNN}} = \frac{T_{\text{full}}}{T_{\text{patch}}^{\text{infer}}} \approx \frac{H \cdot W}{\frac{N}{B} \cdot h \cdot w} = B.
\end{equation}

\subsection{Attention-based Backbone}
For standard attention mechanism, every token attends to all others, leading to a quadratic time complexity with respect to the number of tokens (pixels in our task). 

\subsubsection{(I) Full-image Processing}
Consider an attention-based network where each token attends to all others. Let $t_2$ denote the average processing time per attention computation per token. Then, for an input image of size $H \times W$ flattened into $H \cdot W$ tokens with channel dimension $d$, the processing time is:
\begin{equation}
    T_{\text{full}} = (H \cdot W)^2 \cdot d \cdot t_2. 
    \label{attn_proc}
\end{equation}

Moreover, since the image width $W$, height $H$, and channel dimension $d$ are treated as constants, the time complexity of full-image processing for attention-based backbones becomes:
\begin{equation}
    \mathcal{O}(H^2 W^2).
\end{equation}

\subsubsection{(II) Patch-wise Training}
As the image is divided into $N$ patches, each of size $h \times w$, resulting in $h \cdot w$ tokens per patch. During training, only $M$ out of $N$ patches are selected, with batch sized of $B$ being processed in parallel. The training time becomes:
\begin{equation}
    T_{\text{patch}}^{\text{train}} = \left\lceil \frac{M}{B} \right\rceil \cdot (h \cdot w)^2 \cdot d \cdot t_2.
\end{equation}

Accordingly, the time complexity of attention-based patch-wise training is:
\begin{equation}
    \mathcal{O}(\frac{M}{B} h^2 w^2).
\end{equation}

The theoretical speedup of attention-based patch-wise training is then:
\begin{equation}
    S_{\text{train}}^{\text{Attn}} = \frac{T_{\text{full}}}{T_{\text{patch}}^{\text{train}}} \approx \frac{H^2 W^2}{\frac{M}{B}h^2 w^2}= \frac{N^2}{M}.
\end{equation}

\subsubsection{(III) Patch-wise Inference}
All $N$ patches are used for inference, and they can be grouped into batches of size $B$ for parallel computation. Thus, the inference time is:
\begin{equation}
    T_{\text{patch}}^{\text{infer}} = \left\lceil \frac{N}{B} \right\rceil \cdot (h \cdot w)^2 \cdot d.
\end{equation}

Neglecting constants, the complexity of attention-based patch-wise inference is:
\begin{equation}
    \mathcal{O}( \frac{N}{B} h^2 w^2).
\end{equation}

And the corresponding inference speedup of attention-based patch-wise inference becomes:
\begin{equation}
    S_{\text{infer}}^{\text{Attn}} = \frac{T_{\text{full}}}{T_{\text{patch}}^{\text{infer}}} \approx \frac{H^2 \cdot W^2}{\frac{N}{B} \cdot h^2 \cdot w^2} =N \cdot B.
\end{equation}

\subsection{Summary of Time Complexity and Speedup}
The proposed patch-wise strategy significantly reduces the runtime complexity for both CNN-based and attention-based backbones. During training, it avoids processing the entire image by operating on a small subset of patches and grouping them into mini-batches for parallel updates. During inference, the ability to process all patches in parallel further boosts efficiency.

Table \ref{tab:complexity_summary} summarizes the theoretical time complexity before and after applying the patch-wise strategy across different architectures. Table \ref{tab:patchwise_speedup} further abstracts the theoretical speedup derived from these reductions, showing that attention-based networks, which suffer from quadratic complexity, benefit more significantly.

\begin{table}[h]
\centering
\caption{Comparison of time complexity before and after applying the patch-wise strategy during training and inference.}
\label{tab:complexity_summary}
\renewcommand{\arraystretch}{1.2}
\setlength{\tabcolsep}{6pt}
\begin{tabular}{l|l|c|c}
\toprule
\textbf{Phase} & \textbf{Architecture} & \textbf{Before} & \textbf{After} \\
\midrule
\multirow{2}{*}{Training} 
& CNN-based        & $\mathcal{O}(H W)$             & $\mathcal{O}(\frac{M}{B} h w)$             \\
& Attention-based  & $\mathcal{O}(H^2 W^2)$         & $\mathcal{O}(\frac{M}{B} h^2 w^2)$          \\
\midrule
\multirow{2}{*}{Inference} 
& CNN-based        & $\mathcal{O}(H W)$             & $\mathcal{O}(\frac{N}{B} h w)$              \\
& Attention-based  & $\mathcal{O}(H^2 W^2)$         & $\mathcal{O}(\frac{N}{B} h^2 w^2)$          \\
\bottomrule
\end{tabular}
\end{table}

\begin{table}[h]
\centering
\caption{Theoretical speedup summary of patch-wise strategy across architectures during training and inference.}
\label{tab:patchwise_speedup}
\renewcommand{\arraystretch}{1.2}
\setlength{\tabcolsep}{12pt}
\begin{tabular}{l|c|c}
\toprule
\textbf{Architecture} & \textbf{Training} & \textbf{Inference} \\
\midrule
CNN-based              & $\frac{N}{M}$                   & $B$ \\
Attention-based  & $\frac{N^2}{M}$  & $N \cdot B$ \\
\bottomrule
\end{tabular}
\end{table} 

\section{More Implementation Details}
For datasets, we conduct experiments on four datasets derived from the PanCollection\footnote{\url{https://github.com/liangjiandeng/PanCollection}}. These datasets are captured by four different satellites: WorldView-3 (WV3), WorldView-2 (WV2), QuickBird (QB), and Gaofen-2 (GF2). Among them, WV3 and WV2 provide 8-band multispectral (MS) images, while QB and GF2 provide 4-band MS images. These datasets are selected to demonstrate the robustness of our method across various sensors and different spectral settings.

For patch-wise training and inference, we select 8 patches per image for unsupervised training, with each patch of size $64 \times 64$ in the PAN image. The seed for random selection is set to 1 by \texttt{torch.manual\_seed(1)}. The unsupervised training is conducted over 10 epochs. During inference, each patch is processed with an extra 4-pixel rim to incorporate surrounding context; this rim is removed before stitching to avoid boundary artifacts. A batch size of 32 is employed to enable efficient parallel inference and facilitate effective generalization of the pretrained model to the test data.

For the loss function, we adopt a weighted combination of three terms: $\mathcal L = \eta_1 \mathcal L_{spa} + \eta_2 \mathcal L_{spe} + \eta_3 \mathcal L_{ori}$. Specifically, we set $\eta_1 = 1$, $\eta_2 = 1$, and $\eta_3 = 0.1$. The terms $\mathcal{L}_{spa}$ and $\mathcal{L}_{spe}$ are physics-aware unsupervised losses that enforce spatial and spectral fidelity, respectively, which are critical for high-quality fusion. Therefore, both $\eta_1$ and $\eta_2$ are set to 1 to emphasize their importance. Although maintaining consistency with the original output can prevent overfitting, excessive reliance on it may constrain the model to the original representational capacity of the backbone, thereby hindering generalization. Therefore, $\eta_3$ for original-output consistency is set to a smaller value of 0.1 to strike a proper balance.

For optimization, we employed the Adam optimizer, a robust choice for training deep learning models, with the learning rate set to $1 \times 10^{-4}$ and weight decay at $1 \times 10^{-5}$ to prevent overfitting. The learning rate was carefully tuned to balance convergence speed with stability. We applied the L1 loss criterion for all components of the loss function, which promotes sparsity in the learned features and enhances generalization, particularly when working with high-dimensional data like remote sensing images.

\section{Further Ablation and Discussion}
\subsection{Influence of Number of Selected Patches}
\begin{figure}[!h]
    \centering
    \includegraphics[width=0.8\linewidth]{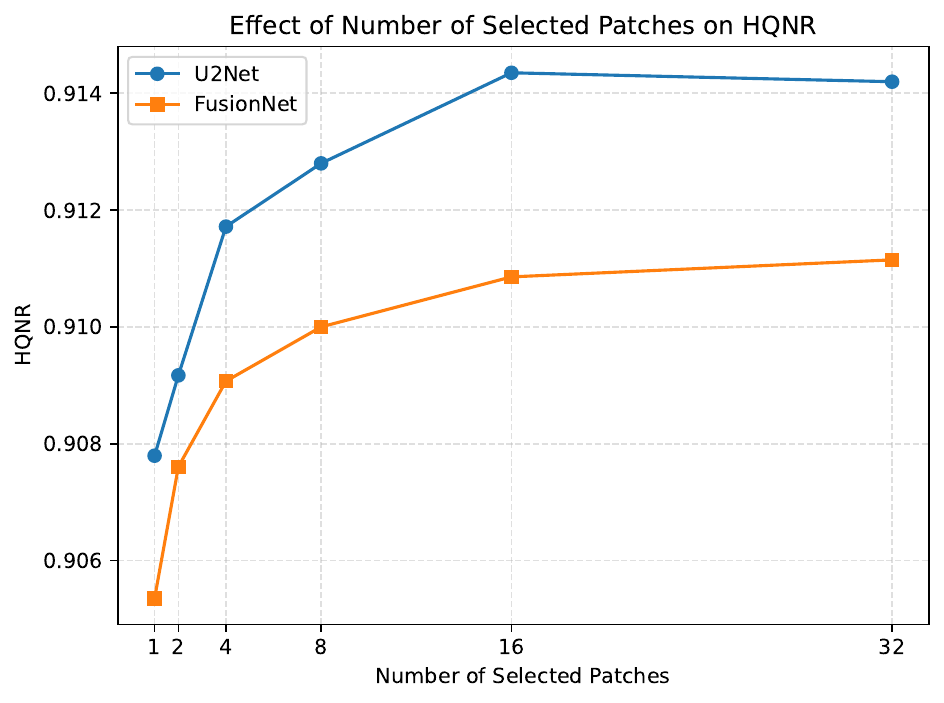}
    \caption{HQNR Metric for U2Net and FusionNet Models with Varying Numbers of Patches on the WV2 Dataset.}
    \label{fig:num_patch}
\end{figure}
In our framework, the number of patches selected for training plays a crucial role in balancing both performance and runtime efficiency. While selecting fewer patches reduces the computational load for both training and inference, it also limits the amount of information available for the model to generalize to the test data. Therefore we conduct experiments with varying number of selected patches from 1 to 32 with FusionNet and U2Net as backbones on WV2 dataset to ensure a comprehensive analysis.

As shown in Figure~\ref{fig:num_patch}, increasing the number of selected patches initially results in significant performance gains. However, the marginal improvement diminishes as more patches are used. Notably, for both models, while performance continues to improve beyond 8 patches, the marginal gains diminish significantly, indicating a rapidly increasing runtime cost for further improvement. This finding indicates that selecting approximately 10\% of the total patches is sufficient to achieve a favorable trade-off, highlighting the efficiency of our patch-wise strategy, enabling strong performance with minimal training cost.

\subsection{Influence of Patch Size}
\begin{figure}[!h]
    \centering
    \includegraphics[width=\linewidth]{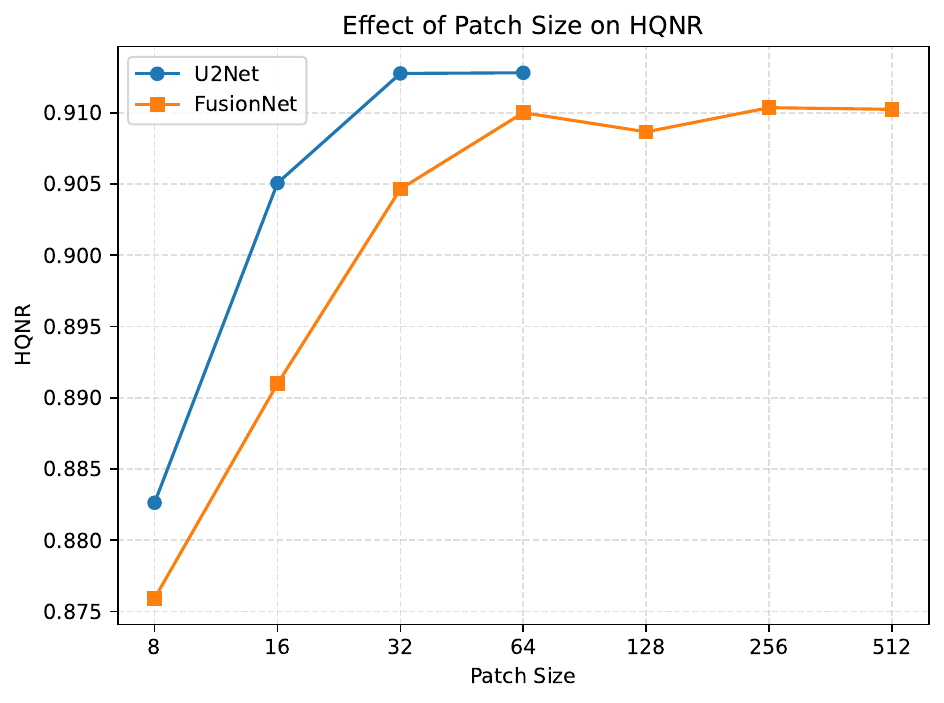}
    \caption{HQNR Metric for U2Net and FusionNet Modelst with Different Patch Sizes on WV2 Dataset.}
    \label{fig:patch_size}
\end{figure}
Patch size is also a crucial factor in balancing performance and computational efficiency. Smaller patches reduce computational complexity but provide less information for the model to generalize to test data, potentially leading to reduced performance. In our experiment, we selected a range of patch sizes within the hardware memory limitations to conduct a comprehensive analysis. The patch sizes ranged from a minimum of $8 \times 8$ to a maximum of $512 \times 512$ in terms of PAN patch size.

As illustrated in Figure~\ref{fig:patch_size}, increasing patch size initially leads to substantial performance gains for both U2Net and FusionNet, with HQNR peaking at $64 \times 64$. Beyond this point, the improvement plateaus, and the trend varies by model: U2Net continues to benefit slightly from larger patches up to the hardware limit, whereas FusionNet experiences a minor decline beyond the peak. These results suggest that larger patches with more contextual information generally facilitate better generalization. Remarkably, a patch size of $64 \times 64$—which is several times smaller than the full $512 \times 512$ PAN image—already yields high performance. This validates the effectiveness of our patch-wise strategy in achieving a favorable balance between efficiency and fusion quality.

\subsection{Influence of Number of Training Epochs}
\begin{figure}[!h]
    \centering
    \includegraphics[width=\linewidth]{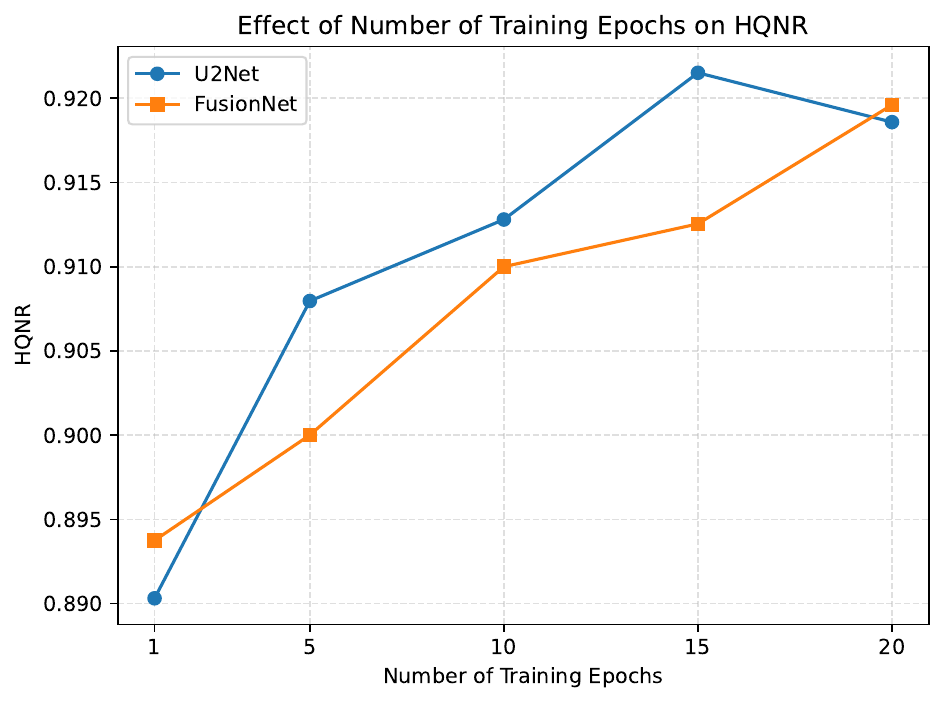}
    \caption{HQNR Metric for U2Net and FusionNet Models with Different Number of Training Epochs on WV2 Dataset.}
    \label{fig:num_epoch}
\end{figure}

\begin{table*}[htbp]
  \centering 
  \setlength{\tabcolsep}{4pt}
  \renewcommand\arraystretch{1.2}
  \resizebox{0.95\linewidth}{!}{
    \begin{tabular}{c|ccc|ccc}
      \toprule
      \toprule
      \multirow{2.2}{*}{Method} & \multicolumn{3}{c|}{WV3 (In-Sensor): Avg$\pm$std} & \multicolumn{3}{c}{WV-2 (Cross-Sensor): Avg$\pm$std} \\
      \cmidrule{2-4}\cmidrule{5-7}
       & HQNR$\uparrow$ & $D_\lambda\downarrow$ & $D_s\downarrow$ & HQNR$\uparrow$ & $D_\lambda\downarrow$ & $D_s\downarrow$ \\
      \midrule
          BT-H         & 0.8659$\pm$0.0568 & 0.0656$\pm$0.0262 & 0.0742$\pm$0.0383 & 0.8300$\pm$0.0430 & 0.0860$\pm$0.0301 & 0.0925$\pm$0.0208 \\
          C-BDSD       & 0.8562$\pm$0.0233 & 0.0874$\pm$0.0236 & 0.0618$\pm$0.0128 & 0.6956$\pm$0.0461 & 0.2253$\pm$0.0488 & 0.1019$\pm$0.0296 \\
          BDSD-PC      & 0.8673$\pm$0.0543 & 0.0634$\pm$0.0246 & 0.0749$\pm$0.0359 & 0.8286$\pm$0.0432 & 0.1413$\pm$0.0320 & \second{0.0356}$\pm$\second{0.0213} \\
          MTF-GLP      & 0.9026$\pm$0.0444 & 0.0373$\pm$0.0124 & 0.0628$\pm$0.0359 & 0.8549$\pm$0.0475 & 0.0582$\pm$0.0221 & 0.0930$\pm$0.0320 \\
          MTF-GLP-FS   & 0.9127$\pm$0.0348 & 0.0357$\pm$0.0106 & 0.0537$\pm$0.0273 & 0.8658$\pm$0.0415 & 0.0563$\pm$0.0212 & 0.0830$\pm$0.0260 \\
          MF           & 0.9014$\pm$0.0358 & 0.0452$\pm$0.0121 & 0.0561$\pm$0.0268 & 0.8508$\pm$0.0538 & 0.0704$\pm$0.0308 & 0.0857$\pm$0.0297 \\
          PsDip        & 0.9215$\pm$0.0176 & 0.0191$\pm$0.0078 & 0.0607$\pm$0.0117 & 0.8980$\pm$0.0226 & \textbf{0.0385}$\pm$\textbf{0.0239} & 0.0659$\pm$0.0158 \\
          ZS-Pan       & 0.9449$\pm$0.0208 & 0.0254$\pm$0.0071 & 0.0306$\pm$0.0153 & \second{0.9112}$\pm$\second{0.0336} & 0.0476$\pm$0.0270 & 0.0435$\pm$0.0130 \\
          FusionMamba  & \second{0.9550}$\pm$\second{0.0110} & 0.0186$\pm$0.0078 & \second{0.0269}$\pm$\second{0.0058} & 0.9022$\pm$0.0242 & 0.0572$\pm$0.0272 & 0.0429$\pm$0.0112 \\
          WFANet       & 0.9459$\pm$0.0093 & \textbf{0.0165}$\pm$\textbf{0.0081} & 0.0382$\pm$0.0041 & \best{0.9128}$\pm$\best{0.0301} & 0.0526$\pm$0.0302 & 0.0366$\pm$0.0054 \\
          FusionNet    & 0.9405$\pm$0.0197 & 0.0229$\pm$0.0080 & 0.0375$\pm$0.0143 & 0.8881$\pm$0.0213 & 0.0543$\pm$0.0273 & 0.0606$\pm$0.0162 \\
          U2Net        & 0.9514$\pm$0.0114 & \second{0.0178}$\pm$\second{0.0072} & 0.0313$\pm$0.0074 & 0.8706$\pm$0.0809 & 0.0936$\pm$0.0790 & 0.0400$\pm$0.0098 \\
          \midrule
          ERFT$_{\text{FusionNet}}$     & 0.9475$\pm$0.0163 & 0.0190$\pm$0.0070 & 0.0342$\pm$0.0120 & 0.9100$\pm$0.0255 & \second{0.0445}$\pm$\second{0.0214} & 0.0475$\pm$0.0160 \\
          ERFT$_{\text{U2Net}}$        & \textbf{0.9572}$\pm$\textbf{0.0132} & 0.0201$\pm$0.0073 & \textbf{0.0232}$\pm$\textbf{0.0092} & \textbf{0.9128}$\pm$\textbf{0.0526} & 0.0617$\pm$0.0494 & \textbf{0.0274}$\pm$\textbf{0.0073} \\
      \bottomrule
      \bottomrule
    \end{tabular}
  }
  \caption{Performance comparison on real-world data of WV3 (in-sensor) and WV2 (cross-sensor) datasets. The reported values represent the average performance over 20 test images. (Best: \textbf{bold}); Second best: \second{underline}}
  \label{tab:wv3_wv2_results}
\end{table*}

The number of training epochs is another key factor influencing the trade-off between performance and computational cost. Fewer epochs result in faster runtime but may lead to underfitting, while more epochs can improve generalization at the cost of increased computation and potential overfitting. To comprehensively assess its impact, we conducted experiments using U2Net and FusionNet on the WV2 dataset, varying the number of unsupervised training epochs from 1 to 20 under a fixed patch-wise setting.

As illustrated in Figure~\ref{fig:num_epoch}, increasing the number of epochs leads to noticeable improvements in HQNR during the initial stages, especially between 1 and 10 epochs. However, the performance gains gradually taper off or become unstable with more training epochs, and also fail to ensure sub-second efficiency with more epochs. These findings indicates that most of the generalization benefits are achieved early in the training process, highlighting that even with a relatively small number of epochs, such as 10, our method can already reach high-quality results, further emphasizing its efficiency and practical applicability.

\subsection{Effectiveness on In-sensor Scenarios}
Although our method is primarily designed to address cross-sensor degradation, achieving substantial quality improvements with sub-second efficiency, we also evaluate its performance in in-sensor scenarios. Specifically, we use models pretrained on the WV3 dataset as the backbone and assess performance on both in-sensor (WV3) and cross-sensor (WV2) test sets. As shown in Table~\ref{tab:reduced_results}, our method continues to outperform its baseline counterparts and achieves state-of-the-art results even in the in-sensor case (WV3), demonstrating its strong robustness and consistently high fusion quality across different scenarios.

\subsection{Influence of Different Loss Terms}
\begin{table}[h]
\centering
\begin{tabular}{lcccc}
\toprule
\textbf{Backbone} & \textbf{Proposed} & {$- \mathcal L_{spa}$} & $- \mathcal L_{spe}$ & $- \mathcal L_{ori}$ \\
\midrule
U2Net      & 0.9128 & 0.9053 & 0.8659 & 0.9131 \\
FusionNet  & 0.9100 & 0.9022 & 0.8963 & 0.9093 \\
\bottomrule
\end{tabular}
\caption{Ablation study of the spatial, spectral, and consistency loss terms ($\mathcal L_{spa}$, $\mathcal L_{spe}$, $\mathcal L_{ori}$) on the WV2 dataset.}
\label{tab:loss_ablation}
\end{table}

Our unsupervised loss function is composed of three components: spatial loss ($\mathcal L_{spa}$), spectral loss ($\mathcal L_{spe}$), and original-output consistency loss ($\mathcal L_{ori}$). To evaluate the contribution of each term, we perform an ablation study by removing them individually. As shown in Table~\ref{tab:loss_ablation}, removing any term generally results in performance degradation, with the spectral loss ($\mathcal L_{spe}$) causing the most significant drop in HQNR. While the impact of removing the consistency loss ($\mathcal L_{ori}$) appears less pronounced in some cases, it still plays a crucial role in preventing overfitting and maintaining robustness, especially on diverse or simulated datasets.

\subsection{Performance on Simulated Data}
\begin{table*}[htbp]
  \centering
  \setlength{\tabcolsep}{4pt}
  \renewcommand\arraystretch{1.2}
  \begin{tabular}{c|cccc}
    \toprule
    \multirow{2}{*}{Method} & \multicolumn{4}{c}{WV3 $\to$ WV2 (Simulated Data): Avg$\pm$std} \\
    \cmidrule{2-5}
    & SAM$\downarrow$ & ERGAS$\downarrow$ & sCC$\uparrow$ & Q8$\uparrow$ \\
    \midrule
    FusionNet                 & 6.3265$\pm$0.6454 & 5.0418$\pm$0.4616 & 0.8813$\pm$0.0136 & 0.7978$\pm$0.0766 \\
    ERFT$_{\text{FusionNet}}$ & {6.3255$\pm$0.6457} & 5.0418$\pm$0.4616 & 0.8813$\pm$0.0136 & 0.7978$\pm$0.0766 \\
    \midrule
    U2Net                     & 5.2586$\pm$0.4991 & 4.0764$\pm$0.3832 & 0.9333$\pm$0.0071 & 0.8470$\pm$0.0846 \\
    ERFT$_{\text{U2Net}}$     & {5.2565$\pm$0.5005} & {4.0739$\pm$0.3839} & 0.9333$\pm$0.0071 & {0.8472$\pm$0.0845} \\
    \bottomrule
  \end{tabular}
  \caption{Performance comparison on simulated data in WV3 $\to$ WV2 case. Metrics are averaged over 20 test images.}
  \label{tab:reduced_results}
\end{table*}
Although our method is primarily designed to enhance pansharpening performance on real-world data, it remains effective when applied to simulated data. Since most loss terms in our framework are tailored for real-world scenarios, and the pretrained model already has an inherent advantage on simulated data (as it was trained with such data), we adjust the loss term weights to $\eta_1 = 10$, $\eta_2 = 100$, and $\eta_3 = 10000$, and initialize the first convolutional layer of the FT module to zero. This configuration emphasizes the original-output consistency loss, helping to preserve the pretrained model’s strengths while still enabling generalization.

As shown in Table \ref{tab:reduced_results}, ERFT-enhanced networks are able to maintain or slightly improve the performance of their baseline counterparts across multiple evaluation metrics. This demonstrates the generalization ability and robustness of the proposed framework, even beyond its original design target.

\subsection{Performance on Newer Architectures}
To validate the generalizability of our proposed method, we extend our evaluation to include newer, state-of-the-art architectures that employ distinct underlying mechanisms. Specifically, we test backbones based on the Mamba mechanism (using FusionMamba) and advanced dense attention models (using WFANet).We test these backbones in two demanding cross-sensor scenarios: $\text{WV-3}\to\text{WV-2}$ and $\text{QB}\to\text{GF-2}$. The performance of the original backbones and their ERFT-enhanced versions ($\text{ERFT}_\text{Backbone}$) are summarized in Table \ref{tab:newer_archs}.

The results demonstrate that our ERFT framework effectively enhances the performance of these modern networks in cross-sensor fusion tasks. Although FusionMamba shows a slight performance drop in the $\text{QB}\to\text{GF-2}$ case, our method consistently boosts the overall fusion quality across different models and scenarios. This effectiveness is particularly evident with WFANet in the $\text{QB}\to\text{GF-2}$ case, where our method achieves a remarkable $\mathbf{11.61\%}$ HQNR rise.

\begin{table}[!ht]

    \centering

    \setlength{\tabcolsep}{6pt}

    \renewcommand\arraystretch{1.2}

    \resizebox{0.9\linewidth}{!}{

    \begin{tabular}{c|l|ccc}

         \toprule

         Case & Backbone & HQNR$\uparrow$ & \(D_\lambda\downarrow\) & \(D_s\downarrow\) \\ 

         \midrule

         \multirow{4}{*}{\shortstack{QB\\$\downarrow $\\GF-2}}

        & FusionMamba   & \best{0.8387} & \second{0.1124} & \best{0.0551} \\ 

        & $\text{ERFT}_\text{FusionMamba}$   & 0.8042 & 0.1200 & \second{0.0860} \\ 

        & WFANet    & 0.7094 & 0.1975 & 0.1154 \\ 

        & $\text{ERFT}_\text{WFANet}$& \second{0.8255} & \best{0.0603} & 0.1211 \\ 

        \midrule

         \multirow{4}{*}{\shortstack{WV-3\\$\downarrow $\\WV-2}}

        & FusionMamba&   0.9022 & 0.0572 & 0.0429 \\ 

        & $\text{ERFT}_\text{FusionMamba}$  & 0.9056 & 0.0563 & 0.0405 \\ 

        & WFANet& \second{0.9128} & \second{0.0526} & \second{0.0366} \\ 

        & $\text{ERFT}_\text{WFANet}$   & \best{0.9360} & \best{0.0379} & \best{0.0272} \\ 

         \bottomrule

    \end{tabular}

    }

    \caption{Performance of newer backbones FusionMamba and WFANet in two cross-sensor cases between our method and corresponding baselines. All values are averaged over test inputs. (\best{Bold}: best; \second{Underline}: second best)}

    \label{tab:newer_archs}

\end{table}

\section{Extensive Application on Megapixel Image}
The application of deep learning-based pansharpening methods on megapixel images (i.e. images containing more than one million pixels) has long been underexplored, primarily due to the substantial computational overhead and prolonged processing times associated with such large-scale data. Traditional deep learning based methods, while effective on data of smaller images, face challenges when applied to high-resolution satellite imagery, where both computational power and time become limiting factors. 

Our proposed pansharpening method addresses these challenges by enabling efficient training and inference on large-scale images in real time. Unlike traditional deep learning approaches, which suffer from long processing times, our method significantly reduces computational complexity, allowing us to perform both training and inference on megapixel images in a matter of seconds. This results in high-quality, superior performance fused images, making it a practical solution for real-time applications in megapixel image pansharpening. Our framework thus opens up new possibilities for fast and efficient processing of large-scale satellite data, offering a scalable and highly effective solution for real-world remote sensing tasks on large-scale data.

\subsection{Data Preparation}
Given the lack of an existing dataset specifically designed for megapixel pansharpening, we constructed a custom dataset using four images from WorldView-3, which were utilized both for pretraining the model and for conducting pansharpening evaluation on megapixel-scale images.
\begin{figure*}[hbtp]
\centering
\includegraphics[width=0.7\linewidth]{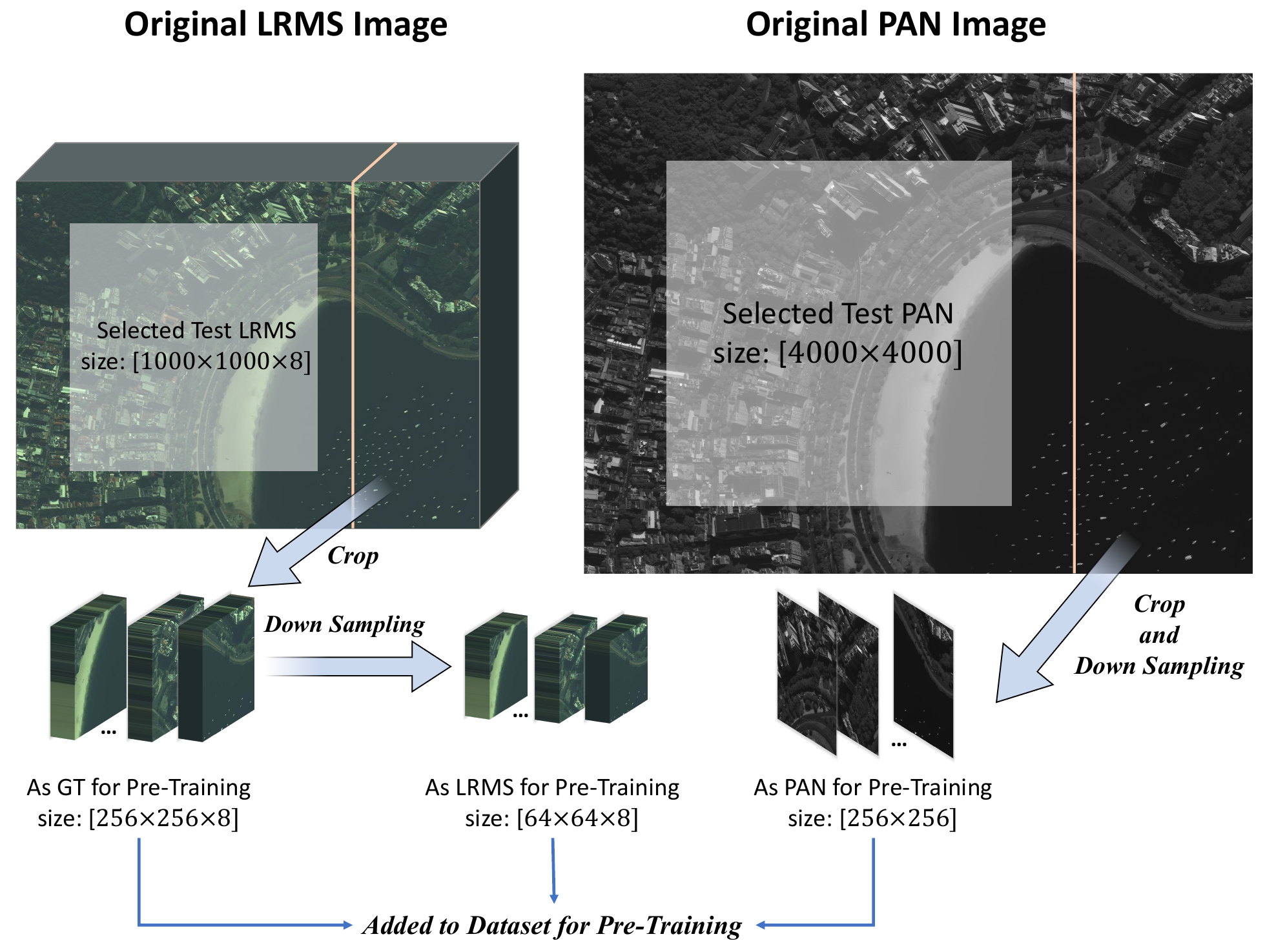}
\caption{Data preparation pipeline for megapixel pansharpening. The pretraining dataset is derived from the right 1/4 portion of the images, while the left 3/4 portion is reserved for test-time evaluation using large-scale megapixel images.}
\label{fig:dataset}
\end{figure*}
For the pretraining phase, we extracted the right 1/4 segment of each image. Specifically, the low-resolution multispectral (LRMS) images were cropped into patches of size $[256 \times 256 \times 8]$, which served as the ground truth (GT) for pretraining. These GT patches were then downsampled to $[64 \times 64 \times 8]$ to serve as the LRMS input data for training. Corresponding panchromatic (PAN) images were cropped into $[1024 \times 1024]$ patches and subsequently downsampled to $[256 \times 256]$ to form the PAN input for pretraining.

For the test-time evaluation, the left 3/4 segment of each image was used. From this portion, we selected a $[1000 \times 1000 \times 8]$ LRMS patch and a corresponding $[4000 \times 4000]$ PAN patch, which were utilized as the test-time input for megapixel image pansharpening.

Using the aforementioned data preparation strategy, we first pre-trained the model on the dataset derived from the right 1/4 segment of the images and subsequently conducted evaluations on the four selected test megapixel images. The results of these experiments are presented in the following subsection.

\subsection{Experiment Results}
\begin{table}[!ht]
    \centering
    \setlength{\tabcolsep}{6pt}
    \renewcommand\arraystretch{1.2}
    \resizebox{1\linewidth}{!}{
    \begin{tabular}{l|ccc|c}
         \toprule
         Method & HQNR$\uparrow$ & $D_\lambda\downarrow$ & $D_s\downarrow$ & Duration$\downarrow$ (s) \\
         \midrule
         BT-H            & 0.7098 & 0.1710 & 0.1444 & \second{2.98} \\
         C-BDSD          & 0.7475 & 0.2019 & 0.0704 & 104.16 \\
         BDSD-PC         & 0.7369 & 0.1723 & 0.1093 & 4.30 \\
         MTF-GLP         & 0.7606 & 0.1231 & 0.1330 & 13.08 \\
         MTF-GLP-FS      & 0.7680 & 0.1272 & 0.1199 & 9.91 \\
         MF              & 0.7709 & 0.1176 & 0.1268 & 7.66 \\
         \midrule
         ERFT$_\text{FusionNet}$ & \second{0.8554} & \textbf{0.0832} & \second{0.0680} & \textbf{2.58} \\
         ERFT$_\text{U2Net}$     & \textbf{0.8590} & \second{0.0972} & \textbf{0.0487} & 22.99 \\
         \bottomrule
    \end{tabular}
    }
    \caption{Performance comparison on megapixel images (PAN size of $4000 \times 4000$). The values represent average performance over 4 test images. (Best: \textbf{bold}; Second best: \second{underline})}
    \label{tab:megapixel_small}
\end{table}
To evaluate the scalability of our method, we conducted experiments on megapixel-scale images, which is an aspect rarely addressed by previous deep learning–based pansharpening methods, to the best of our knowledge. Specifically, we use inputs with a PAN size of $4000 \times 4000$, resulting in HRMS outputs with 128 million pixels. Since conventional deep learning–based pansharpening models cannot directly process such large inputs due to memory and computational constraints, we compare our method against several traditional approaches that rely on mathematical transformations rather than learning-based architectures. 

As shown in Table~\ref{tab:megapixel_small}, traditional methods demonstrate limited performance on large-scale data. In contrast, our method delivers significantly better fusion quality while maintaining a runtime comparable to non-learning-based approaches. These results confirm that deep learning–based pansharpening can scale effectively to megapixel imagery, enabling real-time processing of ultra-high-resolution remote sensing data.

\begin{figure*}
    \centering
    \includegraphics[width=0.95\linewidth]{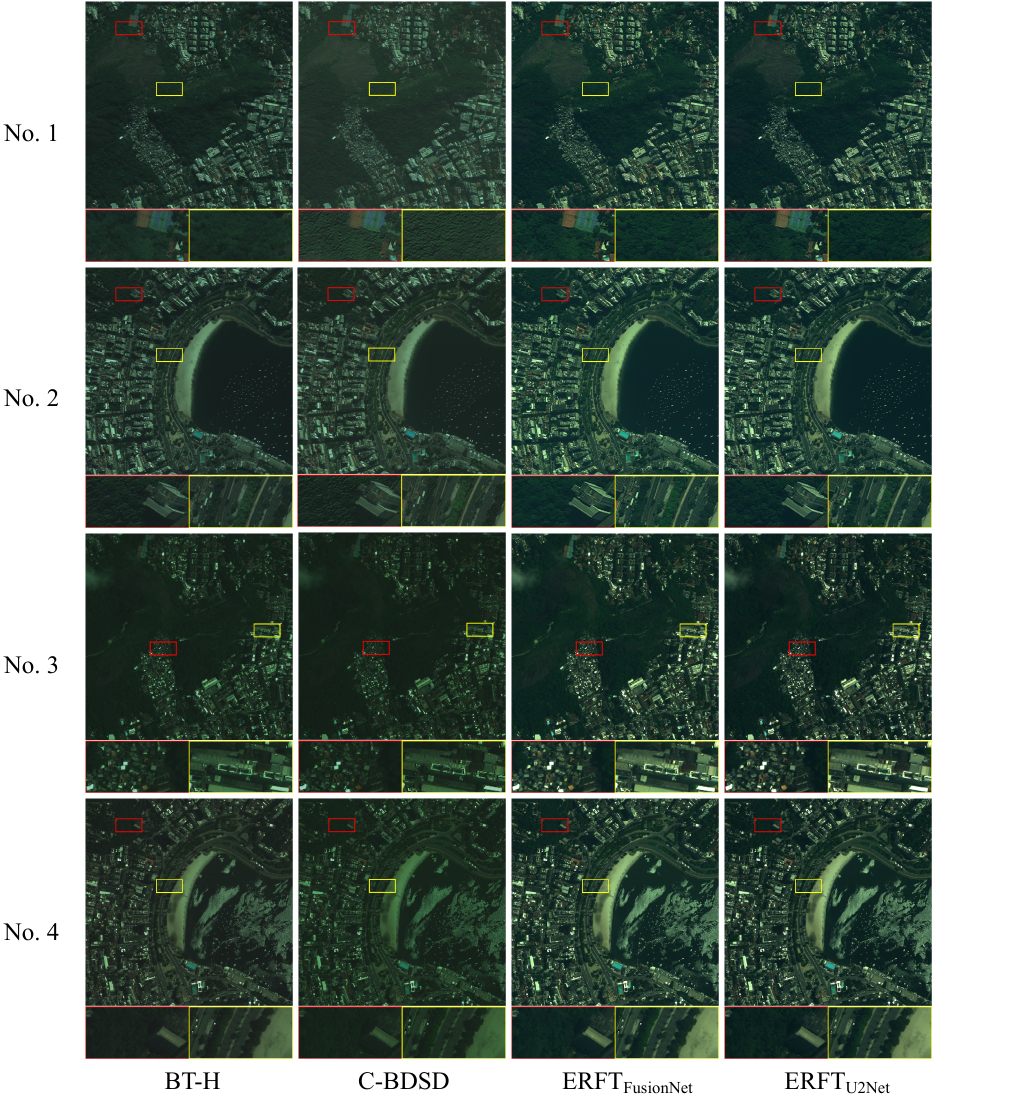}
    \caption{Fusion results on four test images (No. 1–No. 4). Each row corresponds to a different test sample, while each column shows the output from a different method, as labeled below. Red and yellow bounding boxes highlight zoomed-in regions to better illustrate fusion details.}
    \label{fig:big-fig}
\end{figure*}

\begin{figure*}
    \centering
    \includegraphics[width=0.95\linewidth]{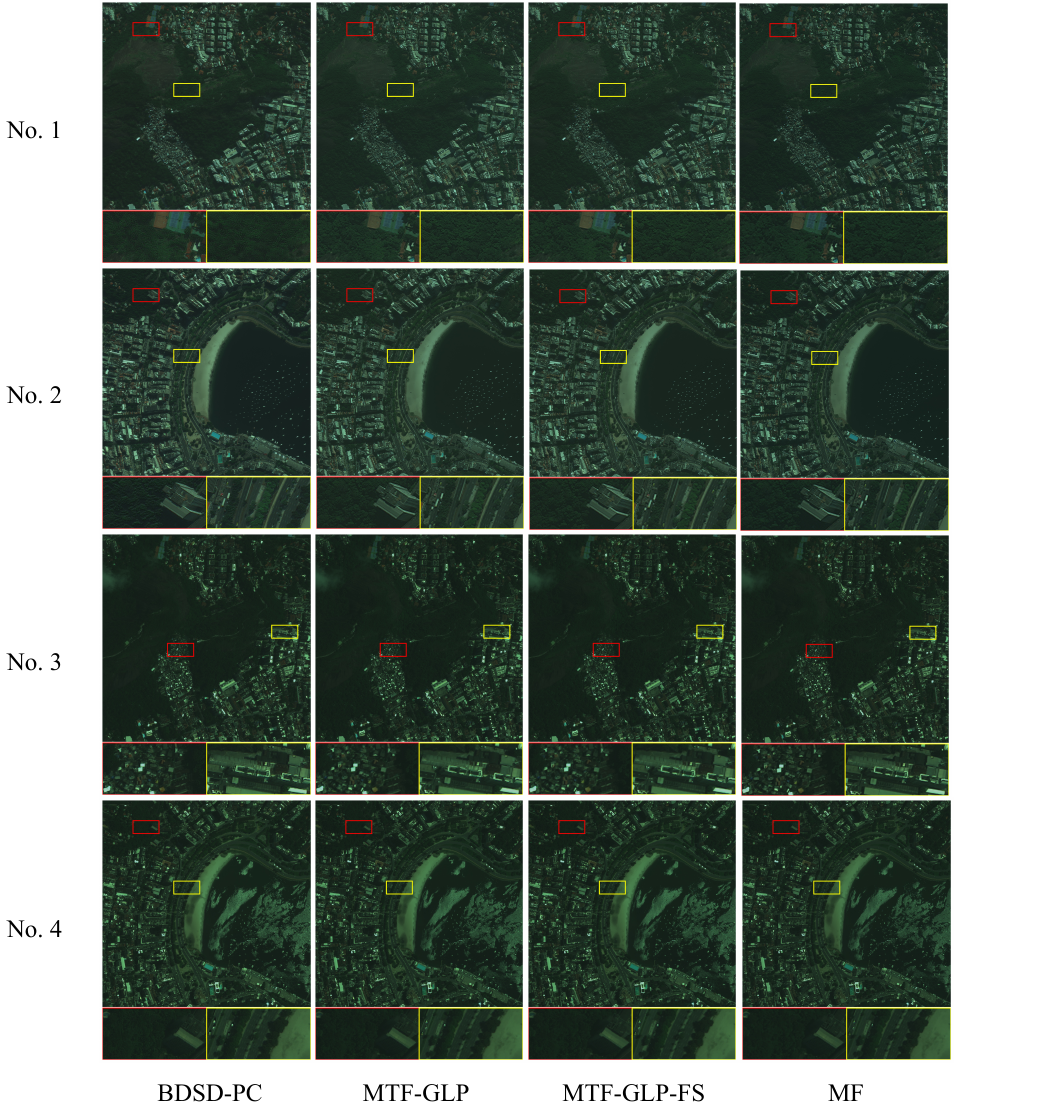}
    \caption{Fusion results on four test images (No. 1–No. 4). Each row corresponds to a different test sample, while each column shows the output from a different method, as labeled below. Red and yellow bounding boxes highlight zoomed-in regions to better illustrate fusion details.}
    \label{fig:big-fig2}
\end{figure*}

Fusion results for four selected test samples obtained using our method, along with the traditional approaches, are presented in Figure \ref{fig:big-fig} and Figure \ref{fig:big-fig2}, where zoomed-in regions are provided to more clearly illustrate the detailed fusion characteristics. Overall, the performance metrics confirm that our approach offers substantial advantages in both the HQNR metric and processing speed, underscoring the practical applicability of our framework for large-scale, real-time pansharpening on megapixel imagery.


\end{document}